%% file: ART_arxiv.tex
\documentclass{article} % For LaTeX2e
\usepackage{iclr2023_conference,times}

\input{math_commands.tex}

\usepackage{url}

\usepackage{booktabs}       % professional-quality tables
\usepackage{amsfonts}       % blackboard math symbols
\usepackage{nicefrac}       % compact symbols for 1/2, etc.
\usepackage{microtype}      % microtypography
\usepackage{xcolor}         % colors

\usepackage{amsmath}
\usepackage{adjustbox}
\usepackage{subfigure}
\usepackage{multirow}
\usepackage{makecell}
\usepackage{epsfig,epstopdf}
\usepackage{bbding}
\usepackage[pagebackref=true,breaklinks=true,colorlinks,bookmarks=false]{hyperref}
\usepackage{wrapfig, blindtext} % put figure inside text
\usepackage{bm} % for bold math

\newcommand\bb[1]{\textbf{#1}}

\usepackage{algpseudocode}
\usepackage[linesnumbered,ruled]{algorithm2e} % for algorithm with vertical lines
\usepackage{arydshln} % table dash line
\title{Accurate Image Restoration with Attention Retractable Transformer}

\author{%
	Jiale Zhang$^{1}$,\enspace Yulun Zhang$^{2}$\thanks{Corresponding authors: Yulun Zhang, yulun100@gmail.com; Linghe Kong, linghe.kong@sjtu.edu.cn},\enspace Jinjin Gu$^{3,4}$,\enspace Yongbing Zhang$^{5}$,\enspace Linghe Kong$^{1 *}$,\enspace Xin Yuan$^{6}$ \\
	\textsuperscript{1}Shanghai Jiao Tong University,\enspace \textsuperscript{2}ETH Z\"{u}rich,\enspace \textsuperscript{3}Shanghai AI Laboratory,\\ \textsuperscript{4}The University of Sydney, \enspace
	\textsuperscript{5}Harbin Institute of Technology (Shenzhen),\enspace \textsuperscript{6}Westlake University%\\
	\vspace{-4mm}
}

\iclrfinalcopy % Uncomment for camera-ready version, but NOT for submission.
\begin{document}

	\maketitle
	
	\begin{abstract}
		Recently, Transformer-based image restoration networks have achieved promising improvements over convolutional neural networks due to parameter-independent global interactions. To lower computational cost, existing works generally limit self-attention computation within non-overlapping windows. However, each group of tokens are always from a dense area of the image. This is considered as a dense attention strategy since the interactions of tokens are restrained in dense regions. Obviously, this strategy could result in restricted receptive fields. To address this issue, we propose \textbf{A}ttention \textbf{R}etractable \textbf{T}ransformer (ART) for image restoration, which presents both dense and sparse attention modules in the network. The sparse attention module allows tokens from sparse areas to interact and thus provides a wider receptive field. Furthermore, the alternating application of dense and sparse attention modules greatly enhances representation ability of Transformer while providing retractable attention on the input image.We conduct extensive experiments on image super-resolution, denoising, and JPEG compression artifact reduction tasks. Experimental results validate that our proposed ART outperforms state-of-the-art methods on various benchmark datasets both quantitatively and visually. We also provide code and models at~\url{https://github.com/gladzhang/ART}.
	\end{abstract}
	
	\setlength{\abovedisplayskip}{2pt}
	\setlength{\belowdisplayskip}{2pt}
	
	%%%%%%%%% BODY TEXT
	\vspace{0mm}
	\section{Introduction}
	Image restoration aims to recover the high-quality image from its low-quality counterpart and includes a series of computer vision applications, such as image super-resolution (SR) and denoising. It is an ill-posed inverse problem since there are a huge amount of candidates for any original input. Recently, deep convolutional neural networks (CNNs) have been investigated to design various models~\cite{kim2016deeply,zhang2020rdnir,zhangASSL} for image restoration. SRCNN~\cite{dong2014learning} firstly introduced deep CNN into image SR. Then several representative works utilized residual learning (e.g., EDSR~\cite{lim2017enhanced}) and attention mechanism (e.g., RCAN~\cite{zhang2018image}) to train very deep network in image SR. Meanwhile, a number of methods were also proposed for image denoising such as DnCNN~\cite{zhang2017beyonddncnn}, RPCNN~\cite{xia2020identifyingRPCNN}, and BRDNet~\cite{tian2020imageBRDNet}. These CNN-based networks have achieved remarkable performance.
	
	However, due to parameter-dependent receptive field scaling and content-independent local interactions of convolutions, CNN has limited ability to model long-range dependencies. To overcome this limitation, recent works have begun to introduce self-attention into computer vision systems~\cite{hu2019local,ramachandran2019stand,wang2020axial,zhao2020exploring}. Since Transformer has been shown to achieve state-of-the-art performance in natural language processing~\cite{vaswani2017attention} and high-level vision tasks~\cite{dosovitskiy2020image,touvron2021training,pvt2021,zheng2021rethinking,chu2021twins}, researchers have been investigating Transformer-based image restoration networks~\cite{yang2020learning,wang2022uformer}. Chen et al. proposed a pre-trained image processing Transformer named IPT~\cite{chen2021preIPT}. Liang et al. proposed a strong baseline model named SwinIR~\cite{swinir2021} based on Swin Transformer~\cite{swintransformer2021} for image restoration. Zamir et al. also proposed an efficient Transformer model using U-net structure named Restormer~\cite{restormer2022} and achieved state-of-the-art results on several image restoration tasks. In contrast, higher performance can be achieved when using Transformer.
	
	\begin{wrapfigure}{r}{0.50\linewidth}
		\centering
		\includegraphics[width=\linewidth]{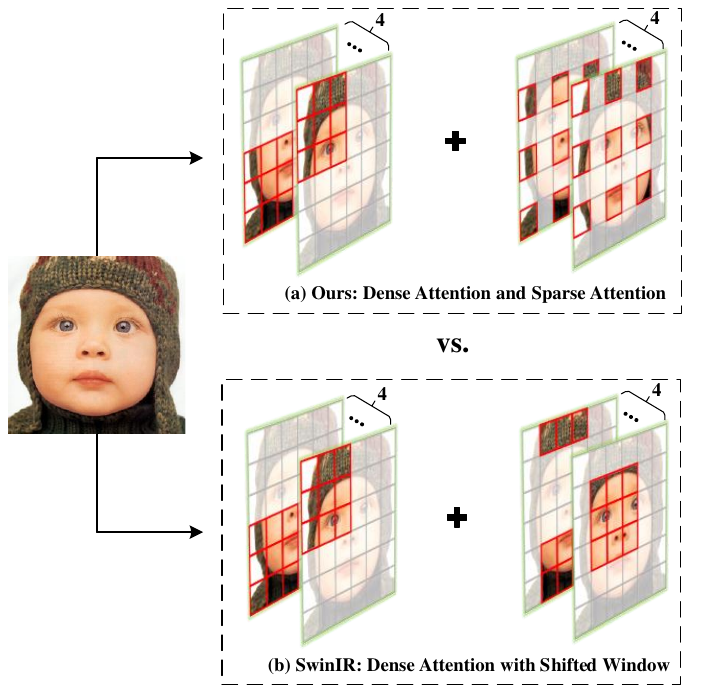}
		\vspace{-7mm}
		\caption{(\textbf{a}) Dense attention and sparse attention strategies of our ART. (\textbf{b}) Dense attention strategy with shifted window of SwinIR.}
		\label{fig:swinvsours}
		\vspace{-3mm}
	\end{wrapfigure}
	Despite showing outstanding performance, existing Transformer backbones for image restoration still suffer from serious defects. As we know, SwinIR~\cite{swinir2021} takes advantage of shifted window scheme to limit self-attention computation within non-overlapping windows. On the other hand, IPT~\cite{chen2021preIPT} directly splits features into $P$$\times$$P$ patches to shrink original feature map $P^2$ times, treating each patch as a token. In short, these methods compute self-attention with shorter token sequences and the tokens in each group are always from a dense area of the image. It is considered as a dense attention strategy, which obviously causes a restricted receptive field. To address this issue, the sparse attention strategy is employed. We extract each group of tokens from a sparse area of the image to provide interactions like previous studies (e.g., GG-Transformer~\cite{gg-transformer2021}, MaxViT~\cite{tu2022maxvit}, CrossFormer~\cite{wang2021crossformer}), but different from them. Our proposed sparse attention module focuses on equal-scale features. Besides, We pay more attention to pixel-level information than semantic-level information. Since the sparse attention has not been well proposed to solve the problems in low-level vision fields, our proposed method can bridge this gap.
	
	We further propose Attention Retractable Transformer named ART for image restoration. Following RCAN~\cite{zhang2018image} and SwinIR~\cite{swinir2021}, we reserve the residual in residual structure~\cite{zhang2018image} for model architecture. Based on joint dense and sparse attention strategies, we design two types of self-attention blocks. We utilize fixed non-overlapping local windows to obtain tokens for the first block named dense attention block (DAB) and sparse grids to obtain tokens for the second block named sparse attention block (SAB). To better understand the difference between our work and SwinIR, we show a visual comparison in Fig.~\ref{fig:swinvsours}. As we can see, the image is divided into four groups and tokens in each group interact with each other. Visibly, the token in our sparse attention block can learn relationships from farther tokens while the one in dense attention block of SwinIR cannot. At the same computational cost, the sparse attention block has stronger ability to compensate for the lack of global information. We consider our dense and sparse attention blocks as successive ones and apply them to extract deep feature. In practice, the alternating application of DAB and SAB can provide retractable attention for the model to capture both local and global receptive field. Our main contributions can be summarized as follows:
	\vspace{-2mm}
	\begin{itemize}
		\item We propose the sparse attention to compensate the defect of mainly using dense attention in existing Transformer-based image restoration networks. The interactions among tokens extracted from a sparse area of an image can bring a wider receptive field to the module.
		\vspace{-0.5mm}
		\item We further propose Attention Retractable Transformer (ART) for image restoration. Our ART offers two types of self-attention blocks to obtain retractable attention on the input feature. With the alternating application of dense and sparse attention blocks, the Transformer model can capture local and global receptive field simultaneously.
		\vspace{-0.5mm}
		\item We employ ART to train an effective Transformer-based network. We conduct extensive experiments on three image restoration tasks: image super-resolution, denoising, and JPEG compression artifact reduction. Our method achieves state-of-the-art performance.
	\end{itemize}
	
	\vspace{-4mm}
	\section{Related Work}
	\vspace{-2mm}
	
	\noindent \textbf{Image Restoration.} With the rapid development of CNN, numerous works based on CNN have been proposed to solve image restoration problems~\cite{anwar2020densely,dudhane2021burst,zamir2020learning,zamir2021multi,li2022blueprint,chen2021attention} and achieved superior performance over conventional restoration approaches~\cite{timofte2013anchored,michaeli2013nonparametric,he2010single}. The pioneering work SRCNN~\cite{dong2014learning} was firstly proposed for image SR. DnCNN~\cite{zhang2017beyonddncnn} was a representative image denoising method. Following these works, various model designs and improving techniques have been introduced into the basic CNN frameworks. These techniques include but not limit to the residual structure~\cite{kim2016accurate,zhang2021plugDRUNet}, skip connection~\cite{zhang2018image,zhang2020rdnir}, dropout \cite{kong2022reflash}, and attention mechanism~\cite{dai2019secondSAN,niu2020singleHAN}. Recently, due to the limited ability of CNN to model long-range dependencies, researchers have started to replace convolution operator with pure self-attention module for image restoration~\cite{yang2020learning,swinir2021,restormer2022,chen2021preIPT}.
	
	\noindent \textbf{Vision Transformer.} Transformer has been achieving impressive performance in machine translation tasks~\cite{vaswani2017attention}. Due to the content-dependent global receptive field, it has been introduced to improve computer vision systems in recent years. Dosovitskiy et al.~\cite{dosovitskiy2020image} proposed ViT and introduced Transformer into image recognition by projecting large image patches into token sequences.  Tu et al. proposed MaxViT~\cite{tu2022maxvit} as an efficient Vision Transformer while introducing multi-axis attention. Wang et al. proposed CrossFormer~\cite{wang2021crossformer} to build the interactions among long and short distance tokens. Yu et al. proposed GG-Transformer~\cite{gg-transformer2021}, which performed self-attention on the adaptively-dilated partitions of the input. Inspired by the strong ability to learn long-range dependencies, researches have also investigated the usage of Transformer for low-level vision tasks~\cite{yang2020learning,chen2021preIPT,swinir2021,restormer2022,wang2022uformer}. However, existing works still suffer from restricted receptive fields due to mainly using dense attention strategy. Very recently, Tu et al. proposed a MLP-based network named MAXIM~\cite{tu2022maxim} to introduce dilated spatial communications into image processing. It further demonstrates that the sparse interactions of visual elements are important for solving low-level problems. In our proposed method, we use dense and sparse attention strategies to build network, which can capture wider global interactions. As the sparse attention has not been well proposed to solve the low-level vision problems, our proposed method can bridge this gap.

	\vspace{-5mm}
	\section{Proposed Method}\label{method}
	\vspace{-3mm}
	% \vspace{-3mm}
	\subsection{Overall Architecture}
\begin{figure*}[t]
		\centering
		\begin{tabular}{c}
			\hspace{-3mm}
			\includegraphics[width=\linewidth]{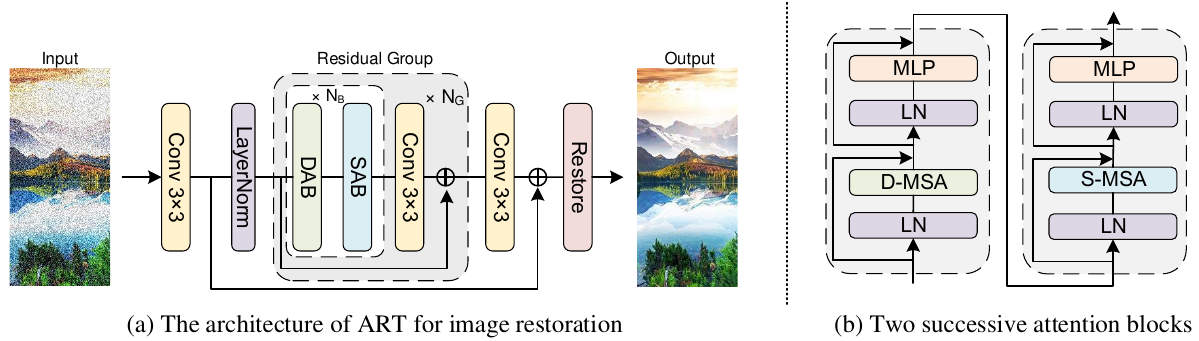} \\
		\end{tabular}
		\vspace{-4mm}
		\caption{(\textbf{a}) The architecture of our proposed ART for image restoration. (\textbf{b}) The inner structure of two successive attention blocks DAB and SAB with two attention modules D-MSA and S-MSA.}
		\label{fig:framework}
		\vspace{-7mm}
	\end{figure*}

	\vspace{-3mm}
	The overall architecture of our ART is shown in Fig.~\ref{fig:framework}. Following RCAN~\cite{zhang2018image}, ART employs residual in residual structure to construct a deep feature extraction module. Given a degraded image $I_{LQ}\in \mathbb{R}^{H\times D \times C_{in}}$ ($H$, $D$, and $C_{in}$ are the height, width, and input channels of the input), ART firstly applies a 3$\times$3 convolutional layer (Conv) to obtain shallow feature $F_0 \in \mathbb{R}^{H\times D \times C}$, where $C$ is the dimension size of the new feature embedding. Next, the shallow feature is normalized and fed into the residual groups, which 
	consist of core Transformer attention blocks. The deep feature is extracted and then passes through another 3$\times$3 Conv to get further feature embeddings $F_1$. Then we use element-wise sum to obtain the final feature map $F_R=F_0 + F_1$. Finally, we employ the restoration module to generate the high-quality image $I_{HQ}$ from the feature map $F_R$.

	\noindent \textbf{Residual Group.} We use $N_G$ successive residual groups to extract the deep feature. Each residual group consists of $N_B$ pairs of attention blocks. We design two successive attention blocks shown in Fig.~\ref{fig:framework}{(b)}. The input feature $x_{l-1}$ passes through layer normalization (LN) and multi-head self-attention (MSA). After adding the shortcut, the output $x{'}_{l}$ is fed into the multi-layer perception (MLP). $x_{l}$ is the final output at the $l$-th block. The process is formulated as
	\begin{equation}
		\label{equ:attention}
		\begin{split}
			x{'}_{l} &= {\rm MSA}({\rm LN}(x_{l-1}))+x_{l-1}, \\
			x_{l} &= {\rm MLP}({\rm LN}(x{'}_{l}))+x{'}_{l}.
		\end{split}
	\end{equation}
	Lastly, we also apply a 3$\times$3 convolutional layer to refine the feature embeddings. As shown in Fig~\ref{fig:framework}(a), a residual connection is employed to obtain the final output in each residual group module.
	
	\noindent \textbf{Restoration Module.} The restoration module is applied as the last stage of the framework to obtain the reconstructed image. As we know, image restoration tasks can be divided into two categories according to the usage of upsampling. For image super-resolution, we take advantage of the sub-pixel convolutional layer~\cite{shi2016real} to upsample final feature map $F_R$. Next, we use a convolutional layer to get the final reconstructed image $I_{HQ}$. The whole process is formulated as
	\begin{equation}
		\label{equ:sr_restore}
		I_{HQ} =  {\rm Conv}({\rm Upsample}(F_R)).
	\end{equation}
	For tasks without upsampling, such as image denoising, we directly use a convolutional layer to reconstruct the high-quality image. Besides, we add the original image to the last output of restoration module for better performance. We formulate the whole process as
	\begin{equation}
		\label{equ:dn_restore}
		I_{HQ} =  {\rm Conv}(F_R) + I_{LQ}.
	\end{equation}
	
	\noindent \textbf{Loss Function.} We optimize our ART with two types of loss functions. There are various well-studied loss functions, such as $L_2$ loss~\cite{dong2016accelerating,sajjadi2017enhancenet,tai2017memnet}, $L_1$ loss~\cite{lai2017deep,zhang2020rdnir}, and Charbonnier loss~\cite{charbonnier1994two}. Same with previous works~\cite{zhang2018image,swinir2021}, we utilize $L_1$ loss for image super-resolution (SR) and Charbonnier loss for image denoising and compression artifact reduction. For image SR, the goal of training ART is to minimize the $L_1$ loss function, which is formulated as
	\begin{equation}
		\label{equ:l1loss}
		\mathcal{L} = \lVert I_{HQ} - I_G\rVert_1,
	\end{equation}
	where $I_{HQ}$ is the output of ART and $I_G$ is the ground-truth image. For image denoising and JPEG compression artifact reduction, we utilize Charbonnier loss with super-parameter $\varepsilon$ as $10^{-3}$, which is
	\begin{equation}
		\label{equ:charbonnierloss}
		\mathcal{L} = \sqrt{\lVert I_{HQ} - I_G\rVert^2 + \varepsilon^2}.
	\end{equation}
	
	\begin{figure*}[t]
		\vspace{-1mm}
		\centering
		\begin{tabular}{c}
			\includegraphics[width=\linewidth]{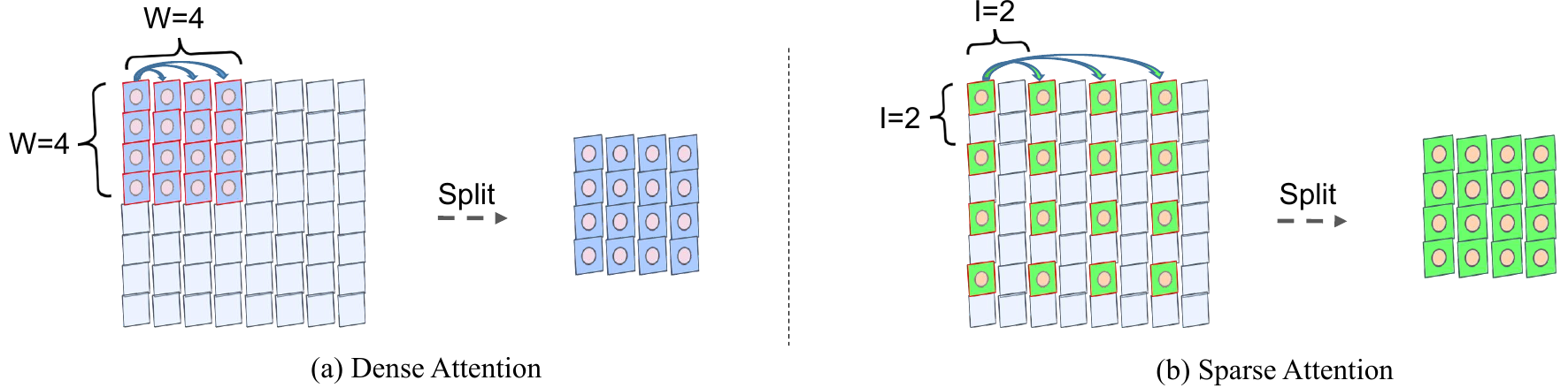} \\
		\end{tabular}
		\vspace{-4mm}
		\caption{(\textbf{a}) Dense attention strategy. Tokens of each group are from a dense area of the image. (\textbf{b}) Sparse attention strategy. Tokens of each group are from a sparse area of the image.}
		\label{fig:attention}
		\vspace{-5.5mm}
	\end{figure*}
	
	\vspace{-3mm}
	\subsection{Attention Retractable Transformer} 
	\label{subsec:art1}
	\vspace{-2mm}
	We elaborate the details about our proposed two types of self-attention blocks in this section. As plotted in Fig.~\ref{fig:framework}{(b)}, the interactions of tokens are concentrated on the multi-head self-attention module (MSA). We formulate the calculation process in MSA as
	\begin{equation}
		\label{equ:msa}
		{\rm MSA}(X) = {\rm Softmax}(\frac{QK^T}{\sqrt{C}})V,
	\end{equation}
	where $Q, K, V \in \mathbb{R}^{N\times C}$ are respectively the query, key, and value from the linear projecting of input $X \in \mathbb{R}^{N\times C}$. $N$ is the length of token sequence, and $C$ is the dimension size of each token. Here we assume that the number of heads is $1$ to transfer MSA to singe-head self-attention for simplification.
	
	\noindent \textbf{Multi-head Self Attention.} Given an image with size $H$$\times$$D$, vision Transformer firstly splits the raw image into numerous patches. These patches are projected by convolutions with stride size $P$. The new projected feature map $\hat{X} \in \mathbb{R}^{h\times w\times C}$ is prepared with $h=\frac{H}{P}$ and $w=\frac{D}{P}$. Common MSA uses all the tokens extracted from the whole feature map and sends them to self-attention module to learn relationships between each other. It will suffer from high computational cost, which is
	\begin{equation}
		\label{equ:msa-complexity}
		\Omega({\rm MSA}) = 4hwC^2 + 2(hw)^2C.
	\end{equation}
	To lower the computational cost, existing works generally utilize non-overlapping windows to obtain shorter token sequences. However, they mainly consider the tokens from a dense area of an image. Different from them, we propose the retractable attention strategies, which provide interactions of tokens from not only dense areas but also sparse areas of an image to obtain a wider receptive field. 
	
	\noindent \textbf{Dense Attention.} As shown in Fig.~\ref{fig:attention}{(a)}, dense attention allows each token to interact with a smaller number of tokens from the neighborhood position of a non-overlapping $W$$\times$$W$ window. All tokens are split into several groups and each group has $W$$\times$$W$ tokens. We apply these groups to compute self-attention for $\frac{h}{W}$$\times$$\frac{w}{W}$ times and the computational cost of new module named D-MSA is
	\begin{equation}
		\label{equ:d-msa-complexity}
		\begin{aligned}
			\Omega({\rm D\mbox{-}MSA}) = (4W^2C^2 + 2W^4C)\times \frac{h}{W}\times \frac{w}{W} = 4hwC^2 + 2W^2hwC.
		\end{aligned}
	\end{equation}
	
	\noindent \textbf{Sparse Attention.} Meanwhile, as shown in Fig.~\ref{fig:attention}{(b)}, we propose sparse attention to allow each token to interact with a smaller number of tokens, which are from sparse positions with interval size $I$. After that, the updates of all tokens are also split into several groups and each group has $\frac{h}{I}$$\times$$\frac{w}{I}$ tokens. We further utilize these groups to compute self-attention for $I$$\times$$I$ times. We name the new multi-head self-attention module as S-MSA and the corresponding computational cost is
	\begin{equation}
		\label{equ:s-msa-complexity}
		\begin{aligned}
			\Omega({\rm S\mbox{-}MSA}) = (4\frac{h}{I}\times \frac{w}{I}C^2 + 2(\frac{h}{I}\times \frac{w}{I})^2C)\times I\times I = 4hwC^2 + 2\frac{h}{I}\frac{w}{I}hwC.
		\end{aligned}
	\end{equation}
	By contrast, our proposed D-MSA and S-MSA modules have lower computational cost since $W^2\ll hw$ and $\frac{h}{I}\frac{w}{I} < hw$. After computing all groups, the outputs are further merged to form original-size feature map. In practice, we apply these two attention strategies to design two types of self-attention blocks named as dense attention block (DAB) and sparse attention block (SAB) as plotted in Fig.~\ref{fig:framework}.
	
	\noindent \textbf{Successive Attention Blocks.} We propose the alternating application of these two blocks. As the local interactions have higher priority, we fix the order of DAB in front of SAB. Besides, we provide the long-distance residual connection between each three pairs of blocks. We show the effectiveness of this joint application with residual connection in the supplementary material.
	
	\noindent \textbf{Attention Retractable Transformer.} We demonstrate that the application of these two blocks enables our model to capture local and global receptive field simultaneously. We treat the successive attention blocks as a whole and get a new type of Transformer named Attention Retractable Transformer, which can provide interactions for both local dense tokens and global sparse tokens.
	
	\begin{table}[t]
		\scriptsize
		\vspace{-4mm}
		\begin{center}
			\begin{tabular}{llllll}
				\hline
				\makecell[l]{Methods}  &\makecell[l]{Solving problems} &\makecell[l]{Structure}&\makecell[l]{Interval of \\ extracted tokens}&\makecell[l]{Representation \\ of tokens}&\makecell[l]{Using long-distance\\residual connection}
				\\ \hline
				GG-Transformer~\cite{gg-transformer2021}       &High-level & Pyramid & Changed & Semantic-level & No \\
				MaxViT~\cite{tu2022maxvit}             &High-level & Pyramid & Changed & Semantic-level & No \\
				CrossFormer~\cite{wang2021crossformer}             &High-level & Pyramid & Changed & Semantic-level & No \\
				ART (Ours)              &Low-level & Isotropic & Unchanged & Pixel-level & Yes \\
				\hline
			\end{tabular}
			\caption{Comparison to related works. The differences between our ART with other works.}
			\label{difference_showing}
		\end{center}
		\vspace{-6mm}
	\end{table}
	\vspace{-2mm}
	\subsection{Differences to related works}
	\vspace{-2mm}
	We summarize the differences between our proposed approach, ART with the closely related works in Tab.~\ref{difference_showing}. We conclude them as three points. \textbf{(1) Different tasks.} GG-Transformer~\cite{gg-transformer2021}, MaxViT~\cite{tu2022maxvit} and CrossFormer~\cite{wang2021crossformer} are proposed to solve high-level vision problems. Our ART is the only one to employ the sparse attention in low-level vision fields. \textbf{(2) Different designs of sparse attention.} In the part of attention, GG-Transformer utilizes the adaptively-dilated partitions, MaxViT utilizes the fixed-size grid attention and CrossFormer utilizes the cross-scale long-distance attention. As the layers get deeper, the interval of tokens from sparse attention becomes smaller and the channels of tokens become larger. Therefore, each token learns more semantic-level information. In contrast, the interval and the channel dimension of tokens in our ART keep unchanged and each token represents the accurate pixel-level information. \textbf{(3) Different model structures.} Different from these works using Pyramid model structure, our proposed ART enjoys an Isotropic structure. Besides, we provide the long-distance residual connection between several Transformer encoders, which enables the feature of deep layers to reserve more low-frequency information from shallow layers. More discussion can be found in the supplementary material.
	
	\vspace{-2mm}
	\subsection{Implementation Details}
	\label{subsec:implementation}
	\vspace{-2mm}
	Some details about how to apply our ART to construct image restoration model are introduced here. Firstly, the residual group number, DAB number, and SAB number in each group are set as $6$, $3$, and $3$. Secondly, all the convolutional layers are equipped with 3$\times$3 kernel, $1$-length stride, and $1$-length padding, so the height and width of feature map remain unchanged. In practice, we treat 1$\times$1 patch as a token. Besides, we set the channel dimension as 180 for most layers except for the shallow feature extraction and the image reconstruction process. Thirdly, the window size in DAB is set as 8 and the interval size in SAB is adjustable according to different tasks, which is discussed in Sec.~\ref{subsec:ablation}. Lastly, to adjust the division of windows and sparse grids, we use padding and mask strategies to the input feature map of self-attention, so that the number of division is always an integer.

	% \vspace{-3mm}
	\section{Experimental Results}
	\vspace{-2mm}
	\subsection{Experimental Settings}
	\label{subsec:settings}
	\vspace{-2mm}
	\noindent \textbf{Data and Evaluation.} We conduct experiments on three image restoration tasks, including image SR, denoising, and JPEG Compression Artifact Reduction (CAR). For image SR, following previous works~\cite{zhang2018image,haris2018deepDBPN}, we use DIV2K~\cite{timofte2017ntire} and Flickr2K~\cite{lim2017enhanced} as training data, Set5~\cite{bevilacqua2012low}, Set14~\cite{zeyde2012single}, B100~\cite{martin2001database}, Urban100~\cite{huang2015single}, and Manga109~\cite{matsui2017sketch} as test data. For image denoising and JPEG CAR, same as SwinIR~\cite{swinir2021}, we use training data: DIV2K, Flickr2K, BSD500~\cite{arbelaez2010contourbsd500}, and WED~\cite{ma2016waterloowed}. We use BSD68~\cite{martin2001database}, Kodak24~\cite{franzen1999kodak}, McMaster~\cite{zhang2011color}, and Urban100 as test data of image denoising. Classic5~\cite{foi2007pointwise} and LIVE1~\cite{sheikh2006statistical} are test data of JPEG CAR. Note that we crop large-size input image into $200$$\times$$200$ partitions with overlapping pixels during inference. Following~\cite{lim2017enhanced}, we adopt the self-ensemble strategy to further improve the performance of our ART and name it as ART+. We evaluate experimental results with PSNR and SSIM~\cite{wang2004image} values on Y channel of images transformed to YCbCr space.
	
	\noindent \textbf{Training Settings.} Data augmentation is performed on the training data through horizontal flip and random rotation of $90^{\circ}$, $180^{\circ}$, and $270^{\circ}$. Besides, we crop the original images into 64$\times$64 patches as the basic training inputs for image SR, 128$\times$128 patches for image denoising, and 126$\times$126 patches for JPEG CAR. We resize the training batch to $32$ for image SR, and $8$ for image denoising and JPEG CAR in order to make a fair comparison. We choose ADAM~\cite{kingma2014adam} to optimize our ART model with $\beta_1=0.9$, $\beta_2=0.999$, and zero weight decay. The initial learning rate is set as 2$\times 10^{-4}$ and is reduced by half as the training iteration reaches a certain number. Taking image SR as an example, we train ART for total $500$k iterations and adjust learning rate to half when training iterations reach $250$k, $400$k, $450$k, and $475$k, where $1$k means one thousand. Our ART is implemented on PyTorch~\cite{paszke2017automatic} with 4 NVIDIA RTX8000 GPUs.

	\begin{figure}[t]
		\centering
		\vspace{-3mm}
		\begin{tabular}{ccc}
			\hspace{-3mm}
			\includegraphics[width=0.34\linewidth]{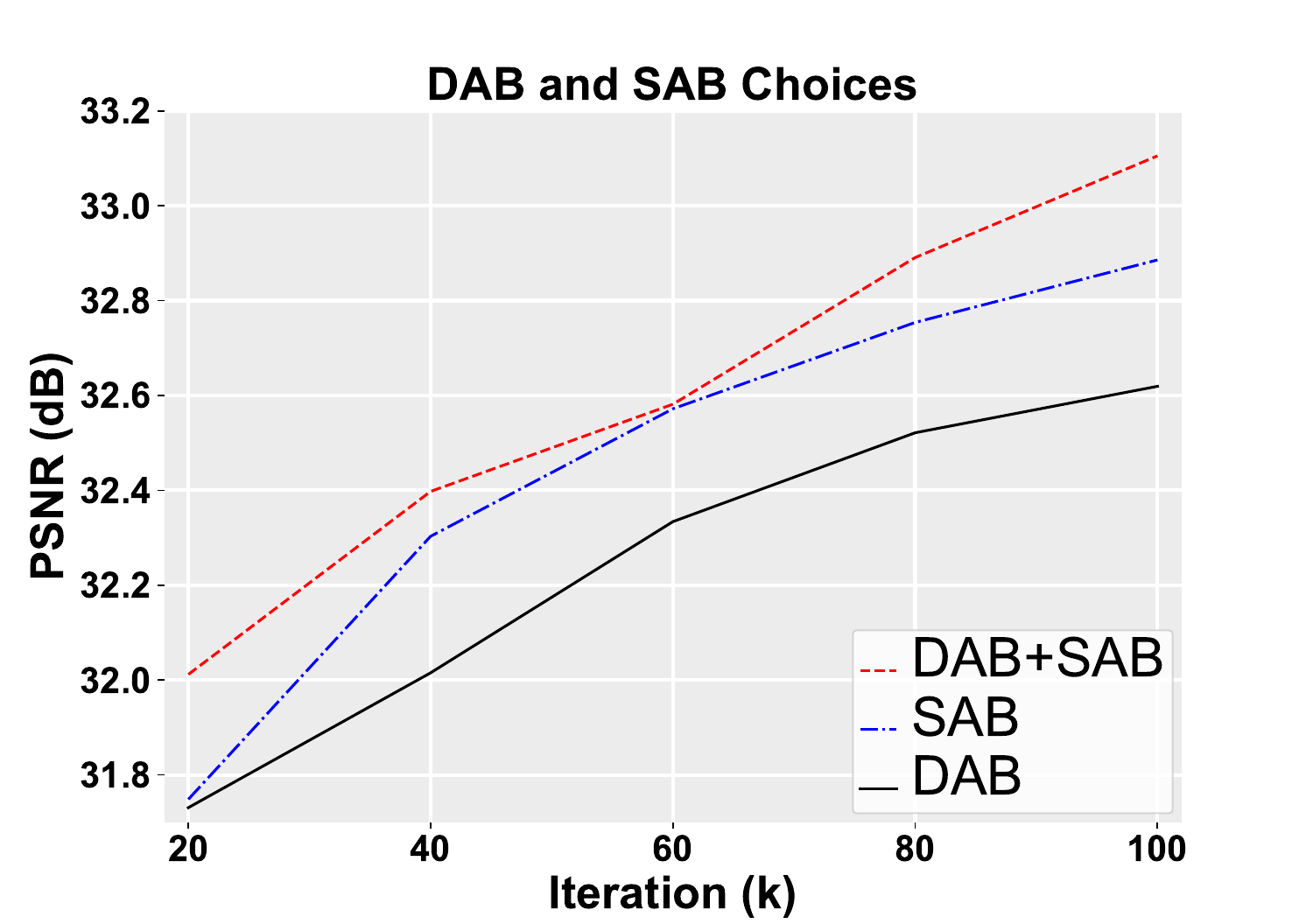} &
			\hspace{-4.5mm}
			\includegraphics[width=0.34\linewidth]{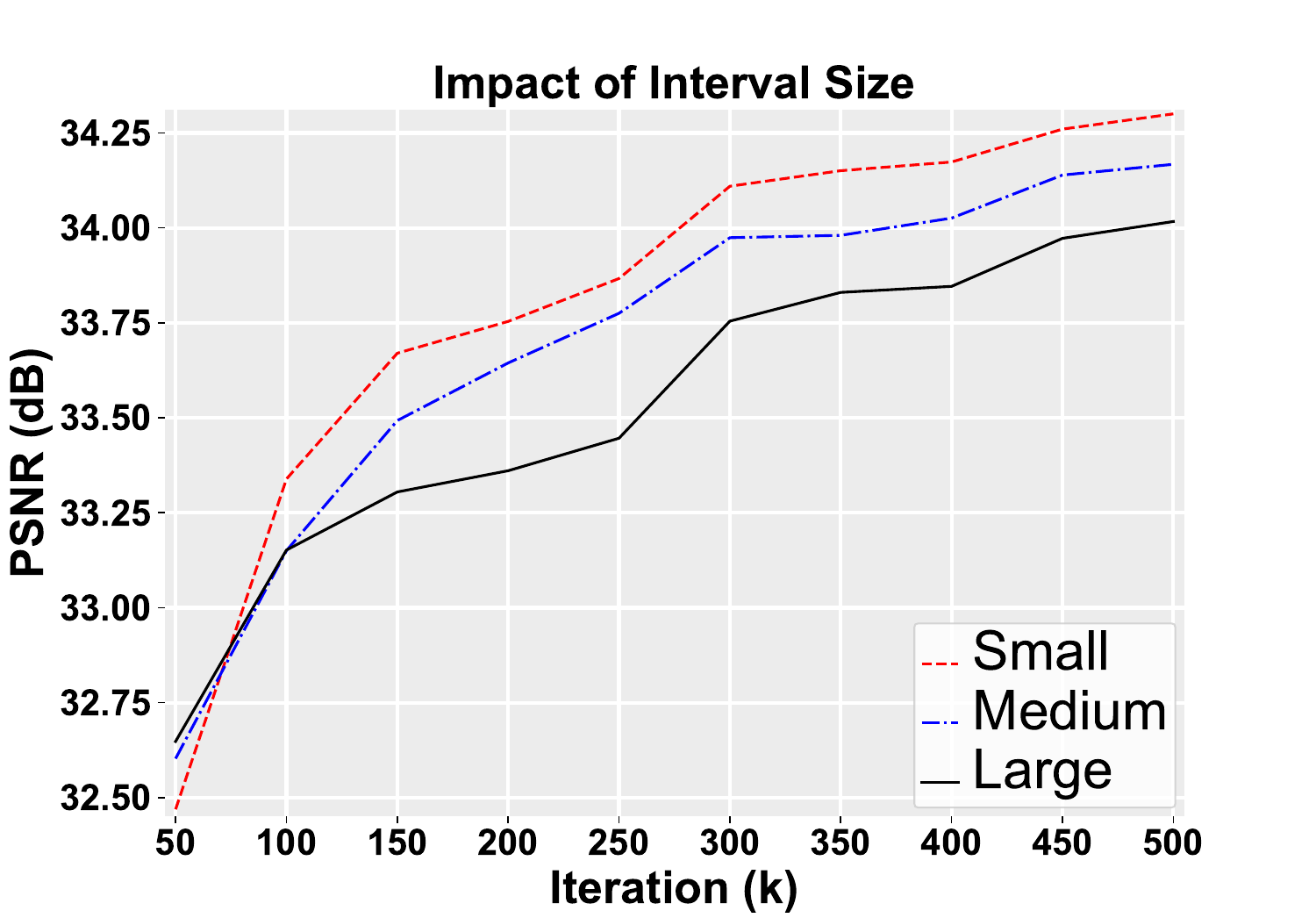} &
			\hspace{-4.5mm}
			\includegraphics[width=0.34\linewidth]{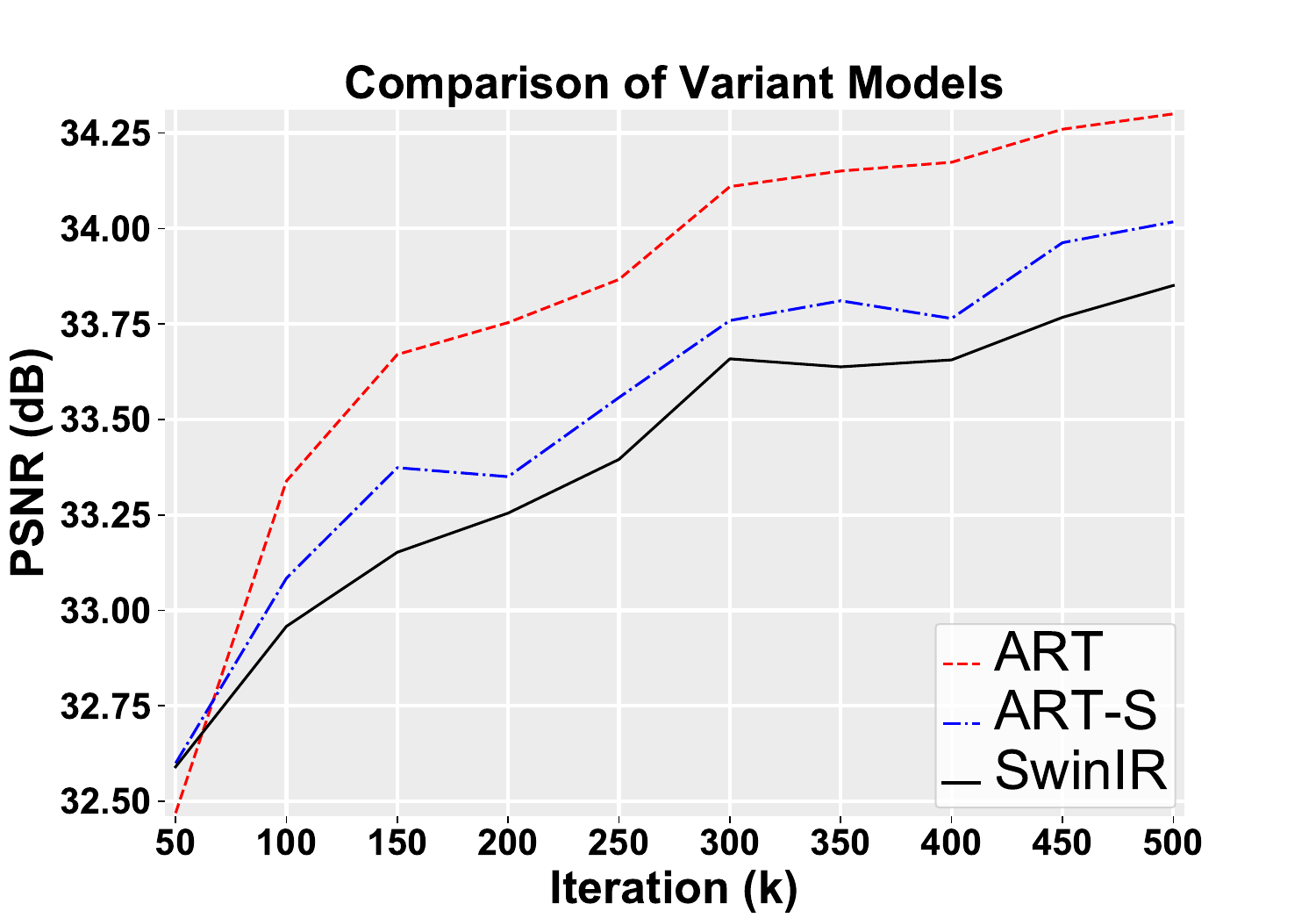} \\
		\end{tabular}
		\vspace{-5mm}
		\caption{\bb{Left:} PSNR (dB) comparison of our ART using all dense attention block (DAB), using all sparse attention block (SAB), and using alternating DAB and SAB. \bb{Middle:} PSNR (dB) comparison of our ART using large interval size in sparse attention block which is $(8,8,8,8,8,8)$ for six residual groups, using medium interval size which is $(8,8,6,6,4,4)$, and using small interval size which is $(4,4,4,4,4,4)$. \bb{Right:} PSNR (dB) comparison of SwinIR, ART-S, and ART.}
		\label{fig:ablation}
		%%\vspace{-1em}
		\vspace{-3.5mm}
	\end{figure}

	\vspace{-2mm}
	\subsection{Ablation Study}
	\label{subsec:ablation}
	\vspace{-2mm}
	For ablation experiments, we train our models for image super-resolution ($\times$2) based on DIV2K and Flicke2K datasets. The results are evaluated on Urban100 benchmark dataset.
	
	\noindent \textbf{Design Choices for DAB and SAB.} We demonstrate the necessity for simultaneous usage of dense attention block (DAB) and sparse attention block (SAB) by conducting ablation study. We set three different experiment conditions, which are using 6 DABs, 6 SABs, and 3 pairs of alternating DAB and SAB. We keep the rest of experiment environment the same and train all models within $100$k iterations. The experimental results are shown in Fig.~\ref{fig:ablation}(Left). As we can see, only using DAB or SAB suffers from poor performance, because they lack either global receptive field or local receptive field. On the other hand, the structure of SAB following DAB brings higher performance. It validates that both local contextual interactions and global sparse interactions are important for improving strong representation ability of Transformer by obtaining retractable attention on the input feature.
	
	\noindent \textbf{Impact of Interval Size.} The interval size in sparse attention block has a vital impact on the performance of our ART. In fact, if the interval size is set as 1, it will be transferred to full attention. Generally, a smaller interval means wider receptive fields but higher computational cost. We compare the experimental results under different interval settings in Fig.~\ref{fig:ablation}(Middle). As we can see, smaller intervals bring more performance gains. To keep the balance between accuracy and complexity, we set the interval size of 6 residual groups as $(4, 4, 4, 4, 4, 4)$ for image SR, $(16,16,12,12,8,8)$ for image denoising, and $(18,18,13,13,7,7)$ for JPEG CAR in the following comparative experiments.
	
	\noindent \textbf{Comparison of Variant Models.} We provide a new version of our model for fair comparisons and name it ART-S. Different from ART, the MLP ratio in ART-S is set to 2 (4 in ART) and the interval size is set to 8. We demonstrate that ART-S has comparable model size with SwinIR. We provide the PSNR comparison results in Fig.~\ref{fig:ablation}(Right). As we can see, our ART-S achieves better performance than SwinIR. More comparative results can be found in following experiment parts.
	
	\begin{table*}[t]
		% \scriptsize
		\tiny
		\begin{center}
			\vspace{-3mm}
			\begin{tabular}{|l|c|c|c|c|c|c|c|c|c|c|c|}
				\hline
				\multirow{2}{*}{Method} & \multirow{2}{*}{Scale} &  \multicolumn{2}{c|}{Set5} & \multicolumn{2}{c|}{Set14} & \multicolumn{2}{c|}{B100} & \multicolumn{2}{c|}{Urban100} & \multicolumn{2}{c|}{Manga109}\\
				\cline{3-12}
				& & PSNR & SSIM & PSNR & SSIM & PSNR & SSIM & PSNR & SSIM & PSNR & SSIM
				\\
				\hline
				\hline
				EDSR~\cite{lim2017enhanced} & $\times$2 & 
				38.11 & 0.9602 & 33.92 & 0.9195 & 32.32 & 0.9013 & 32.93 & 0.9351 & 39.10 & 0.9773\\
				RCAN~\cite{zhang2018image} & $\times$2 &
				38.27 & 0.9614 & 34.12 & 0.9216 & 32.41 & 0.9027 & 33.34 & 0.9384 & 39.44 & 0.9786\\
				SAN~\cite{dai2019secondSAN} & $\times$2 &
				38.31 & 0.9620 & 34.07 & 0.9213 & 32.42 & 0.9028 & 33.10 & 0.9370 & 39.32 & 0.9792\\
				SRFBN~\cite{li2019feedbackSRFBN} & $\times$2 &
				38.11 & 0.9609 & 33.82 & 0.9196 & 32.29 & 0.9010 & 32.62 & 0.9328 & 39.08 & 0.9779\\
				HAN~\cite{niu2020singleHAN} & $\times$2 &
				38.27 & 0.9614 & 34.16 & 0.9217 & 32.41 & 0.9027 & 33.35 & 0.9385 & 39.46 & 0.9785\\
				IGNN~\cite{zhou2020crossIGNN} & $\times$2 &
				38.24 & 0.9613 & 34.07 & 0.9217 & 32.41 & 0.9025 & 33.23 & 0.9383 & 39.35 & 0.9786\\
				CSNLN~\cite{mei2020imageCSNLN} & $\times$2 &
				38.28 & 0.9616 & 34.12 & 0.9223 & 32.40 & 0.9024 & 33.25 & 0.9386 & 39.37 & 0.9785\\
				RFANet~\cite{liu2020residualRFANet} & $\times$2 &
				38.26 & 0.9615 & 34.16 & 0.9220 & 32.41 & 0.9026 & 33.33 & 0.9389 & 39.44 & 0.9783\\
				NLSA~\cite{mei2021imageNLSA} & $\times$2 &
				38.34 & 0.9618 & 34.08 & 0.9231 & 32.43 & 0.9027 & 33.42 & 0.9394 & 39.59 & 0.9789\\
				IPT~\cite{chen2021preIPT} & $\times$2 &
				38.37 & N/A & 34.43 & N/A & 32.48 & N/A & 33.76 & N/A & N/A & N/A \\
				SwinIR~\cite{swinir2021} & $\times$2 &
				38.42 & 0.9623 & 34.46 & 0.9250 & 32.53 & 0.9041 & 33.81 & 0.9427 & 39.92 & 0.9797\\
				\textbf{ART-S (ours)} & $\times$2 &
				38.48 & 0.9625 & 34.50 & 0.9258 & 32.53 & 0.9043 & 34.02 & 0.9437 & 40.11 & 0.9804\\
				
				\textbf{ART (ours)} & $\times$2 &  
				\textcolor{blue}{38.56} & \textcolor{blue}{0.9629} & \textcolor{blue}{34.59} & \textcolor{blue}{0.9267} & \textcolor{blue}{32.58} & \textcolor{blue}{0.9048} & \textcolor{blue}{34.30} & \textcolor{blue}{0.9452} & \textcolor{blue}{40.24} & \textcolor{blue}{0.9808}
				\\ 
				\textbf{ART+ (ours)} & $\times$2 &  
				\textcolor{red}{38.59} & \textcolor{red}{0.9630} & \textcolor{red}{34.68} & \textcolor{red}{0.9269} & \textcolor{red}{32.60} & \textcolor{red}{0.9050} & \textcolor{red}{34.41} & \textcolor{red}{0.9457} & \textcolor{red}{40.33} & \textcolor{red}{0.9810}
				\\
				
				\hline
				\hline
				EDSR~\cite{lim2017enhanced} & $\times$3 & 
				34.65 & 0.9280 & 30.52 & 0.8462 & 29.25 & 0.8093 & 28.80 & 0.8653 & 34.17 & 0.9476\\
				RCAN~\cite{zhang2018image} & $\times$3 &
				34.74 & 0.9299 & 30.65 & 0.8482 & 29.32 & 0.8111 & 29.09 & 0.8702 & 34.44 & 0.9499\\
				SAN~\cite{dai2019secondSAN} & $\times$3 &
				34.75 & 0.9300 & 30.59 & 0.8476 & 29.33 & 0.8112 & 28.93 & 0.8671 & 34.30 & 0.9494\\
				SRFBN~\cite{li2019feedbackSRFBN} & $\times$3 &
				34.70 & 0.9292 & 30.51 & 0.8461 & 29.24 & 0.8084 & 28.73 & 0.8641 & 34.18 & 0.9481\\
				HAN~\cite{niu2020singleHAN} & $\times$3 &
				34.75 & 0.9299 & 30.67 & 0.8483 & 29.32 & 0.8110 & 29.10 & 0.8705 & 34.48 & 0.9500\\
				IGNN~\cite{zhou2020crossIGNN} & $\times$3 &
				34.72 & 0.9298 & 30.66 & 0.8484 & 29.31 & 0.8105 & 29.03 & 0.8696 & 34.39 & 0.9496\\
				CSNLN~\cite{mei2020imageCSNLN} & $\times$3 &
				34.74 & 0.9300 & 30.66 & 0.8482 & 29.33 & 0.8105 & 29.13 & 0.8712 & 34.45 & 0.9502\\
				RFANet~\cite{liu2020residualRFANet} & $\times$3 &
				34.79 & 0.9300 & 30.67 & 0.8487 & 29.34 & 0.8115 & 29.15 & 0.8720 & 34.59 & 0.9506\\
				NLSA~\cite{mei2021imageNLSA} & $\times$3 &
				34.85 & 0.9306 & 30.70 & 0.8485 & 29.34 & 0.8117 & 29.25 & 0.8726 & 34.57 & 0.9508\\
				IPT~\cite{chen2021preIPT} & $\times$3 &
				34.81 & N/A & 30.85 & N/A & 29.38 & N/A & 29.49 & N/A & N/A & N/A \\
				SwinIR~\cite{swinir2021} & $\times$3 &
				34.97 & 0.9318 & 30.93 & 0.8534 & 29.46 & 0.8145 & 29.75 & 0.8826 & 35.12 & 0.9537\\
				\textbf{ART-S (ours)} & $\times$3 &
				34.98 & 0.9318 & 30.94 & 0.8530 & 29.45 & 0.8146 & 29.86 & 0.8830 & 35.22 & 0.9539\\
				
				\textbf{ART (ours)} & $\times$3 &  
				\textcolor{blue}{35.07} & \textcolor{blue}{0.9325} & \textcolor{blue}{31.02} & \textcolor{blue}{0.8541} & \textcolor{blue}{29.51} & \textcolor{blue}{0.8159} & \textcolor{blue}{30.10} & \textcolor{blue}{0.8871} & \textcolor{blue}{35.39} & \textcolor{blue}{0.9548}
				\\ 
				\textbf{ART+ (ours)} & $\times$3 &  
				\textcolor{red}{35.11} & \textcolor{red}{0.9327} & \textcolor{red}{31.05} & \textcolor{red}{0.8545} & \textcolor{red}{29.53} & \textcolor{red}{0.8162} & \textcolor{red}{30.22} & \textcolor{red}{0.8883} & \textcolor{red}{35.51} & \textcolor{red}{0.9552}
				\\
				% \\
				\hline
				\hline
				EDSR~\cite{lim2017enhanced} & $\times$4 & 
				32.46 & 0.8968 & 28.80 & 0.7876 & 27.71 & 0.7420 & 26.64 & 0.8033 & 31.02 & 0.9148\\
				RCAN~\cite{zhang2018image} & $\times$4 &
				32.63 & 0.9002 & 28.87 & 0.7889 & 27.77 & 0.7436 & 26.82 & 0.8087 & 31.22 & 0.9173\\
				SAN~\cite{dai2019secondSAN} & $\times$4 &
				32.64 & 0.9003 & 28.92 & 0.7888 & 27.78 & 0.7436 & 26.79 & 0.8068 & 31.18 & 0.9169\\
				SRFBN~\cite{li2019feedbackSRFBN} & $\times$4 &
				32.47 & 0.8983 & 28.81 & 0.7868 & 27.72 & 0.7409 & 26.60 & 0.8015 & 31.15 & 0.9160\\
				HAN~\cite{niu2020singleHAN} & $\times$4 &
				32.64 & 0.9002 & 28.90 & 0.7890 & 27.80 & 0.7442 & 26.85 & 0.8094 & 31.42 & 0.9177\\
				IGNN~\cite{zhou2020crossIGNN} & $\times$4 &
				32.57 & 0.8998 & 28.85 & 0.7891 & 27.77 & 0.7434 & 26.84 & 0.8090 & 31.28 & 0.9182\\
				CSNLN~\cite{mei2020imageCSNLN} & $\times$4 &
				32.68 & 0.9004 & 28.95 & 0.7888 & 27.80 & 0.7439 & 27.22 & 0.8168 & 31.43 & 0.9201\\
				RFANet~\cite{liu2020residualRFANet} & $\times$4 &
				32.66 & 0.9004 & 28.88 & 0.7894 & 27.79 & 0.7442 & 26.92 & 0.8112 & 31.41 & 0.9187\\
				NLSA~\cite{mei2021imageNLSA} & $\times$4 &
				32.59 & 0.9000 & 28.87 & 0.7891 & 27.78 & 0.7444 & 26.96 & 0.8109 & 31.27 & 0.9184\\
				IPT~\cite{chen2021preIPT} & $\times$4 &
				32.64 & N/A & 29.01 & N/A & 27.82 & N/A & 27.26 & N/A & N/A & N/A \\
				SwinIR~\cite{swinir2021} & $\times$4 &
				32.92 & 0.9044 & 29.09 & 0.7950 & 27.92 & 0.7489 & 27.45 & 0.8254 & 32.03 & 0.9260\\
				\textbf{ART-S (ours)} & $\times$4 &
				32.86 & 0.9029 & 29.09 & 0.7942 & 27.91 & 0.7489 & 27.54 & 0.8261 & 32.13 & 0.9263\\
				
				\textbf{ART (ours)} & $\times$4 &  
				\textcolor{blue}{33.04} & \textcolor{blue}{0.9051} & \textcolor{blue}{29.16} & \textcolor{blue}{0.7958} & \textcolor{blue}{27.97} & \textcolor{blue}{0.7510} & \textcolor{blue}{27.77} & \textcolor{blue}{0.8321} & \textcolor{blue}{32.31} & \textcolor{blue}{0.9283}
				\\ 
				\textbf{ART+ (ours)} & $\times$4 &  
				\textcolor{red}{33.07} & \textcolor{red}{0.9055} & \textcolor{red}{29.20} & \textcolor{red}{0.7964} & \textcolor{red}{27.99} & \textcolor{red}{0.7513} & \textcolor{red}{27.89} & \textcolor{red}{0.8339} & \textcolor{red}{32.45} & \textcolor{red}{0.9291}
				\\
				\hline
			\end{tabular}
			\vspace{-2mm}
			\caption{PSNR (dB)/SSIM comparisons for image super-resolution on five benchmark datasets. We color best and second best results in \textcolor{red}{red} and \textcolor{blue}{blue}.}
			\label{table:psnr_ssim_SR_5sets}
		\end{center}
		\vspace{-1mm}
	\end{table*}

	\begin{table*}[t]
		\scriptsize
		%\small
		\vspace{-2mm}
		\begin{center}
			\begin{tabular}{|l|c|c|c|c|c|c|c|c|}
				\hline
				Method & EDSR & RCAN & SRFBN & HAN & CSNLN & SwiIR & ART-S (ours) & ART (ours)\\
				\hline
				\hline
				Params (M) & 43.09 & 15.59 & 3.63 & 16.07 & 7.16 & 11.90 & 11.87 & 16.55\\
				\hline
				Mult-Adds (G) & 1,286 & 407 & 498 & 420 & 103,640 & 336 & 392 & 782\\
				\hline
				PSNR on Urban100 (dB) & 26.64 & 26.82 & 26.60 & 26.85 & 27.22 & 27.45 & 27.54 & 27.77\\
				\hline
				PSNR on Manga109 (dB) & 31.02 & 31.22 & 31.15 & 31.42 & 31.43 & 32.03 & 32.13 & 32.31\\
				\hline
			\end{tabular}
			\vspace{-2mm}
			\caption{Model size comparisons ($\times$4 SR). Output size is 3$\times$640$\times$640 for Mult-Adds calculation.}
			\label{table:model_size}
		\end{center}
		\vspace{-8mm}
	\end{table*}
	
	\begin{figure*}[t]
		\tiny
		\centering
		\begin{tabular}{cc}
			% % one row
			\hspace{-0.42cm}
			\begin{adjustbox}{valign=t}
				\begin{tabular}{c}
					\includegraphics[width=0.213\textwidth]{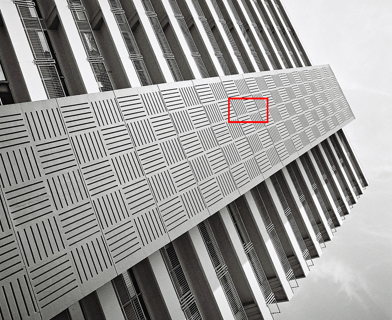}
					\\
					Urban100: img\_092 ($\times$4)
				\end{tabular}
			\end{adjustbox}
			\hspace{-0.46cm}
			\begin{adjustbox}{valign=t}
				\begin{tabular}{cccccc}
					\includegraphics[width=0.149\textwidth]{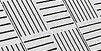} \hspace{-4mm} &
					\includegraphics[width=0.149\textwidth]{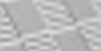} \hspace{-4mm} &
					\includegraphics[width=0.149\textwidth]{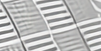} \hspace{-4mm} &
					\includegraphics[width=0.149\textwidth]{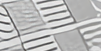} \hspace{-4mm} &
					\includegraphics[width=0.149\textwidth]{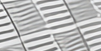} \hspace{-4mm} 
					\\
					HQ / PSNR (dB) \hspace{-4mm} &
					Bicubic / 15.31\hspace{-4mm} &
					RCAN / 18.36 \hspace{-4mm} &
					SRFBN / 18.26 \hspace{-4mm} &
					SAN / 18.26\hspace{-4mm}
					\\
					\includegraphics[width=0.149\textwidth]{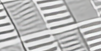} \hspace{-4mm} &
					\includegraphics[width=0.149\textwidth]{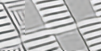} \hspace{-4mm} &
					\includegraphics[width=0.149\textwidth]{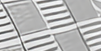} \hspace{-4mm} &
					\includegraphics[width=0.149\textwidth]{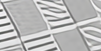} \hspace{-4mm} &
					\includegraphics[width=0.149\textwidth]{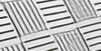} \hspace{-4mm}  
					\\ 
					IGNN / 18.51 \hspace{-4mm} &
					CSNLN / 18.69 \hspace{-4mm} &
					RFANet / 18.49 \hspace{-4mm} &
					SwinIR / 18.59 \hspace{-4mm} &
					\textbf{ART / 19.56} \hspace{-4mm}
					\\
				\end{tabular}
			\end{adjustbox}
			\\
			% % one row
			\hspace{-0.42cm}
			\begin{adjustbox}{valign=t}
				\begin{tabular}{c}
					\includegraphics[width=0.213\textwidth]{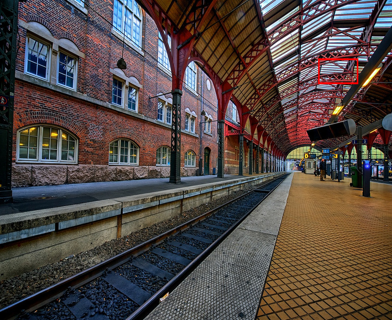}
					\\
					Urban100: img\_098 ($\times$4)
				\end{tabular}
			\end{adjustbox}
			\hspace{-0.46cm}
			\begin{adjustbox}{valign=t}
				\begin{tabular}{cccccc}
					\includegraphics[width=0.149\textwidth]{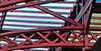} \hspace{-4mm} &
					\includegraphics[width=0.149\textwidth]{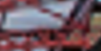} \hspace{-4mm} &
					\includegraphics[width=0.149\textwidth]{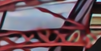} \hspace{-4mm} &
					\includegraphics[width=0.149\textwidth]{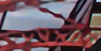} \hspace{-4mm} &
					\includegraphics[width=0.149\textwidth]{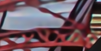} \hspace{-4mm} 
					\\
					HQ / PSNR (dB) \hspace{-4mm} &
					Bicubic / 18.28\hspace{-4mm} &
					RCAN / 19.70 \hspace{-4mm} &
					SRFBN / 19.55 \hspace{-4mm} &
					SAN / 19.66 \hspace{-4mm}
					\\
					\includegraphics[width=0.149\textwidth]{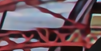} \hspace{-4mm} &
					\includegraphics[width=0.149\textwidth]{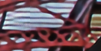} \hspace{-4mm} &
					\includegraphics[width=0.149\textwidth]{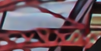} \hspace{-4mm} &
					\includegraphics[width=0.149\textwidth]{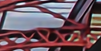} \hspace{-4mm} &
					\includegraphics[width=0.149\textwidth]{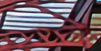} \hspace{-4mm}  
					\\ 
					IGNN / 19.70 \hspace{-4mm} &
					CSNLN / 19.82 \hspace{-4mm} &
					RFANet / 19.72 \hspace{-4mm} &
					SwinIR / 20.00 \hspace{-4mm} &
					\textbf{ART / 20.10} \hspace{-4mm}
					\\
				\end{tabular}
			\end{adjustbox}
			
		\end{tabular}
		\vspace{-3.5mm}
		\caption{Visual comparison with challenging examples on image super-resolution ($\times$4).}
		\label{fig:img_sr_visual}
		\vspace{-7mm}
	\end{figure*}

	\vspace{-2.5mm}
	\subsection{Image Super-Resolution}
	\vspace{-2mm}
	We provide comparisons of our proposed ART with representative image SR methods, including CNN-based networks: EDSR~\cite{lim2017enhanced}, RCAN~\cite{zhang2018image}, SAN~\cite{dai2019secondSAN},
	SRFBN~\cite{li2019feedbackSRFBN}, HAN~\cite{niu2020singleHAN}, IGNN~\cite{zhou2020crossIGNN}, CSNLN~\cite{mei2020imageCSNLN}, RFANet~\cite{liu2020residualRFANet}, NLSA~\cite{mei2021imageNLSA}, and Transformer-based networks: IPT~\cite{chen2021preIPT} and SwinIR~\cite{swinir2021}. Note that IPT is a pre-trained model, which is trained on ImageNet benchmark dataset. All the results are provided by publicly available code and data. Quantitative and visual comparisons are provided in Tab.~\ref{table:psnr_ssim_SR_5sets} and Fig.~\ref{fig:img_sr_visual}.
	
	\noindent \textbf{Quantitative Comparisons.} We present PSNR/SSIM comparison results for $\times$2, $\times$3, and $\times$4  image SR in Tab.~\ref{table:psnr_ssim_SR_5sets}. As we can see, our ART achieves the best PSNR/SSIM performance on all five benchmark datasets. Using self-ensemble, ART+ gains even better results. Compared with existing state-of-the-art method SwinIR, our ART obtains better gains across all scale factors, indicating that our proposed joint dense and sparse attention blocks enable Transformer stronger representation ability. Despite showing better performance than CNN-based networks, another Transformer-based network IPT is not as good as ours. It is validated that our proposed ART becomes a new promising Transformer-based network for image SR.
	
	\begin{wrapfigure}{r}{0.55\linewidth}
		\centering
		\vspace{-4mm}
		% \rule{0.9\linewidth}{0.75\linewidth}
		\includegraphics[width=\linewidth]{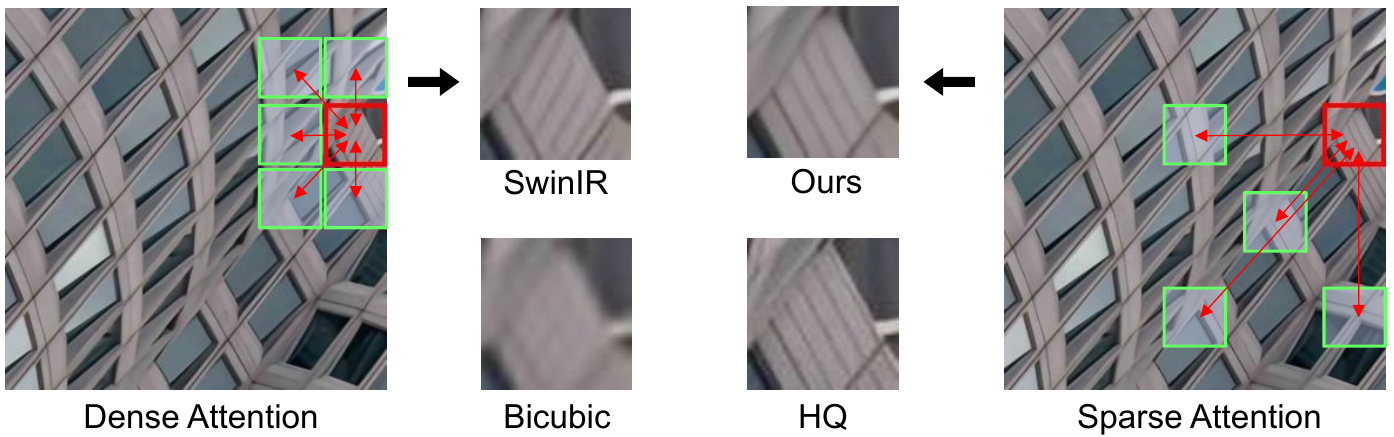}
		\vspace{-8mm}
		\caption{Visual comparison ($\times$4) of SwinIR and Ours.}
		\label{fig:visual_comp}
		\vspace{-3mm}
	\end{wrapfigure}
	\noindent \textbf{Retractable vs. Dense Attention.} We further show a typical visual comparison with SwinIR in Fig.~\ref{fig:visual_comp}. As SwinIR mainly utilizes dense attention strategy, it restores wrong texture structures under the influence of close patches with mainly vertical lines. However, our ART can reconstruct the right texture, thanks to the wider receptive field provided by sparse attention strategy. Visibly, the patch is able to interact with farther patches with similar horizontal lines so that it can be reconstructed clearly. This comparison demonstrates the advantage of retractable attention and its strong ability to restore high-quality outputs.
	
	\noindent \textbf{Model Size Comparisons.} Table~\ref{table:model_size} provides comparisons of parameters number and Mult-Adds of different networks, which include existing state-of-the-art methods. We calculate the Mult-Adds assuming that the output size is 3$\times$640$\times$640 under $\times$4 image SR. Compared with previous CNN-based networks, our ART has comparable parameter number and Mult-Adds but achieves high performance. Besides, we can see that our ART-S has less parameters and Mult-Adds than most of the compared methods. The model size of ART-S is similar with SwinIR. However, ART-S still achieves better performance gains than all compared methods except our ART. It indicates that our method is able to achieve promising performance at an acceptable computational and memory cost. 
	
	\noindent \textbf{Visual Comparisons.} We also provide some challenging examples for visual comparison ($\times$4) in Fig.~\ref{fig:img_sr_visual}. We can see that our ART is able to alleviate heavy blurring artifacts while restoring detailed edges and textures. Compared with other methods, ART obtains visually pleasing results by recovering more high-frequency details. It indicates that ART preforms better for image SR.

	\vspace{-3mm}
	\subsection{Image Denoising}
\begin{figure*}[ht]
		\tiny
		\centering
		\vspace{-2mm}
		\begin{tabular}{cc}
			% % one row
			\hspace{-0.45cm}
			\begin{adjustbox}{valign=t}
				\begin{tabular}{c}
					\includegraphics[width=0.213\textwidth]{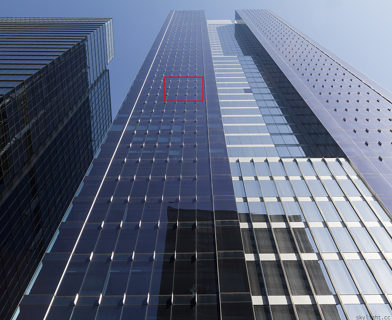}
					\\
					Urban100: img\_033
				\end{tabular}
			\end{adjustbox}
			\hspace{-0.46cm}
			\begin{adjustbox}{valign=t}
				\begin{tabular}{cccccc}
					\includegraphics[width=0.149\textwidth]{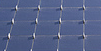} \hspace{-4mm} &
					\includegraphics[width=0.149\textwidth]{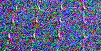} \hspace{-4mm} &
					\includegraphics[width=0.149\textwidth]{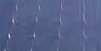} \hspace{-4mm} &
					\includegraphics[width=0.149\textwidth]{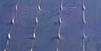} \hspace{-4mm} &
					\includegraphics[width=0.149\textwidth]{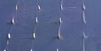}
					\hspace{-4mm} 
					\\
					HQ / PSNR (dB) \hspace{-4mm} &
					Noisy / 15.15 \hspace{-4mm} &
					CBM3D / 28.72 \hspace{-4mm} &
					IRCNN / 28.57 \hspace{-4mm} &
					DnCNN / 29.13 \hspace{-4mm}
					\\
					\includegraphics[width=0.149\textwidth]{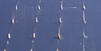} \hspace{-4mm} &
					\includegraphics[width=0.149\textwidth]{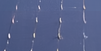} \hspace{-4mm} &
					\includegraphics[width=0.149\textwidth]{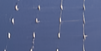} \hspace{-4mm} &
					\includegraphics[width=0.149\textwidth]{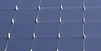} \hspace{-4mm} &
					\includegraphics[width=0.149\textwidth]{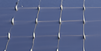} \hspace{-4mm}  
					\\ 
					RNAN / 30.28 \hspace{-4mm} &
					RDN / 30.48 \hspace{-4mm} &
					SwinIR / 31.16  \hspace{-4mm} &
					Restormer / 31.81 \hspace{-4mm} &
					\textbf{ART / 31.94 }\hspace{-4mm}
					\\
				\end{tabular}
			\end{adjustbox}
		\end{tabular}
		\vspace{-3.5mm}
		\caption{Visual comparison with challenging examples on color image denoising ($\sigma$=50).}
		\label{fig:img_dn_visual}
		\vspace{-4mm}
	\end{figure*}

	\vspace{-2mm}
	We show color image denoising results to compare our ART with representative methods in Tab.~\ref{table:psnr_DN_4sets}. These methods are CBM3D~\cite{Dabov2007CBM3D}, IRCNN~\cite{zhang2017learningIRCNN}, FFDNet~\cite{zhang2018ffdnet}, DnCNN~\cite{zhang2017beyonddncnn}, RNAN~\cite{zhang2019rnan}, RDN~\cite{zhang2020rdnir}, IPT~\cite{chen2021preIPT}, DRUNet~\cite{zhang2021plugDRUNet}, P3AN~\cite{hu2021pseudo}, SwinIR~\cite{swinir2021}, and Restormer~\cite{restormer2022}. Following most recent works, we set the noise level to $15$, $25$, and $50$. We also shows visual comparisons of challenging examples in Fig.~\ref{fig:img_dn_visual}.
	
	\noindent \textbf{Quantitative Comparisons.} Table~\ref{table:psnr_DN_4sets} shows PSNR results of color image denoising. As we can see, our ART achieves the highest performance across all compared methods on three datasets except Kodak24. Even better results are obtained by ART+ using self-ensemble. Particularly, it obtains better gains than the state-of-the-art model Restormer~\cite{restormer2022} by up to 0.25dB on Urban100. Restormer also has restricted receptive fields and thus has difficulty in some challenging cases. In conclusion, these comparisons indicate that our ART also has strong ability in image denoising. 
	
	\begin{table*}[ht]
		% \scriptsize
		\tiny
		\begin{center}
			\resizebox{\textwidth}{!}{
				\setlength{\tabcolsep}{0.8mm}
				\vspace{-5mm}
				\begin{tabular}{|l|c|c|c|c|c|c|c|c|c|c|c|c|}
					\hline
					\multirow{2}{*}{Method} &  \multicolumn{3}{c|}{BSD68} & \multicolumn{3}{c|}{Kodak24} & \multicolumn{3}{c|}{McMaster} & \multicolumn{3}{c|}{Urban100}\\
					\cline{2-13}
					&$\sigma$=15 & $\sigma$=25 & $\sigma$=50 &$\sigma$=15 & $\sigma$=25 & $\sigma$=50 &$\sigma$=15 & $\sigma$=25 & $\sigma$=50 &$\sigma$=15 & $\sigma$=25 & $\sigma$=50\\
					\hline
					\hline
					CBM3D~\cite{Dabov2007CBM3D} &
					N/A & N/A & 27.38 & N/A & N/A & 28.63 & N/A & N/A & N/A & N/A & N/A & 27.94\\
					IRCNN~\cite{zhang2017learningIRCNN} &
					33.86 & 31.16 & 27.86 & 34.69 & 32.18 & 28.93 & 34.58 & 32.18 & 28.91 & 33.78 & 31.20 & 27.70\\
					FFDNet~\cite{zhang2018ffdnet} &
					33.87 & 31.21 & 27.96 & 34.63 & 32.13 & 28.98 & 34.66 & 32.35 & 29.18 & 33.83 & 31.40 & 28.05\\
					DnCNN~\cite{zhang2017beyonddncnn} &
					33.90 & 31.24 & 27.95 & 34.60 & 32.14 & 28.95 & 33.45 & 31.52 & 28.62 & 32.98 & 30.81 & 27.59\\
					RNAN~\cite{zhang2019rnan} &
					N/A & N/A & 28.27 & N/A & N/A & 29.58 & N/A & N/A & 29.72 & N/A & N/A & 29.08\\
					RDN~\cite{zhang2020rdnir} &
					N/A & N/A & 28.31 & N/A & N/A & 29.66 & N/A & N/A & N/A & N/A & N/A & 29.38 \\
					IPT~\cite{chen2021preIPT} &
					N/A & N/A & 28.39 & N/A & N/A & 29.64 & N/A & N/A & 29.98 & N/A & N/A & 29.71\\
					DRUNet~\cite{zhang2021plugDRUNet} &
					34.30 & 31.69 & 28.51 & 35.31 & 32.89 & \textcolor{black}{29.86} & 35.40 & 33.14 & 30.08 & 34.81 & 32.60 & 29.61\\
					P3AN~\cite{hu2021pseudo} &
					N/A & N/A & 28.37 & N/A & N/A & 29.69 & N/A & N/A & N/A & N/A & N/A & 29.51\\
					SwinIR~\cite{swinir2021} &
					34.42 & 31.78 & 28.56 & 35.34 & 32.89 & 29.79 & 35.61 & 33.20 & 30.22 & 35.13 & 32.90 & 29.82\\
					Restormer~\cite{restormer2022} &
					34.40 & 31.79 & \textcolor{black}{28.60} & \textcolor{red}{35.47} & \textcolor{red}{33.04} & \textcolor{red}{30.01} & 35.61 & 33.34 & \textcolor{black}{30.30} & 35.13 & 32.96 & 30.02\\
					
					\textbf{ART (ours)} &  
					\textcolor{blue}{34.46} & \textcolor{blue}{31.84} & \textcolor{blue}{28.63} & \textcolor{black}{35.39} & \textcolor{black}{32.95} & 29.87 & \textcolor{blue}{35.68} & \textcolor{blue}{33.41} & \textcolor{blue}{30.31} & \textcolor{blue}{35.29} & \textcolor{blue}{33.14} & \textcolor{blue}{30.19}
					\\
					\textbf{ART+ (ours)} &
					\textcolor{red}{34.47} & \textcolor{red}{31.85} & \textcolor{red}{28.65} & \textcolor{blue}{35.41} & \textcolor{blue}{32.98} & \textcolor{blue}{29.89} & \textcolor{red}{35.71} & \textcolor{red}{33.44} & \textcolor{red}{30.35} & \textcolor{red}{35.34} & \textcolor{red}{33.20} & \textcolor{red}{30.27}
					\\
					\hline
			\end{tabular} }
			\vspace{-4mm}
			\caption{PSNR (dB) comparisons. The best and second best results are in \textcolor{red}{red} and \textcolor{blue}{blue}.}
			\label{table:psnr_DN_4sets}
		\end{center}
		\vspace{-2mm}
	\end{table*}
	
	\begin{table*}[t]
		\scriptsize
		\begin{center}
			\vspace{-3mm}
			\resizebox{\textwidth}{!}{
				\setlength{\tabcolsep}{1.5mm}
				\begin{tabular}{|l|c|c|c|c|c|c|c|c|c|c|c|c|c|}
					\hline
					\multirow{2}{*}{Dataset} & \multirow{2}{*}{$q$} &  \multicolumn{2}{c|}{RNAN} & \multicolumn{2}{c|}{RDN} & \multicolumn{2}{c|}{DRUNet} & \multicolumn{2}{c|}{SwinIR} & 
					\multicolumn{2}{c|}{ART (ours)} &
					\multicolumn{2}{c|}{ART+ (ours)}
					\\
					\cline{3-14}
					
					& & PSNR & SSIM & PSNR & SSIM & PSNR & SSIM & PSNR & SSIM & PSNR & SSIM & PSNR & SSIM
					\\
					\hline
					\hline
					
					\multirow{3}{*}{Classic5}
					& 10
					& 29.96 & 0.8178
					& 30.00 & 0.8188
					& 30.16 & 0.8234
					& \textcolor{black}{30.27} & \textcolor{black}{0.8249}
					& \textcolor{blue}{30.27} & \textcolor{blue}{0.8258}
					& \textcolor{red}{30.32} & \textcolor{red}{0.8263}
					\\
					& 30
					& 33.38 & 0.8924
					& 33.43 & 0.8930
					& 33.59 & 0.8949
					& \textcolor{black}{33.73} & \textcolor{black}{0.8961}
					& \textcolor{blue}{33.74} & \textcolor{blue}{0.8964}
					& \textcolor{red}{33.78} & \textcolor{red}{0.8967}
					\\
					& 40
					& 34.27 & 0.9061
					& 34.27 & 0.9061
					& 34.41 & 0.9075
					& \textcolor{black}{34.52} & \textcolor{black}{0.9082}
					& \textcolor{blue}{34.55} & \textcolor{blue}{0.9086}
					& \textcolor{red}{34.58} & \textcolor{red}{0.9089}
					\\
					\hline
					\multirow{3}{*}{LIVE1}
					& 10
					& 29.63 & 0.8239
					& 29.67 & 0.8247
					& 29.79 & 0.8278
					& \textcolor{black}{29.86} & \textcolor{black}{0.8287}
					& \textcolor{blue}{29.89} & \textcolor{blue}{0.8300}
					& \textcolor{red}{29.92} & \textcolor{red}{0.8305}
					\\
					& 30
					& 33.45 & 0.9149
					& 33.51 & 0.9153
					& 33.59 & 0.9166
					& \textcolor{black}{33.69} & \textcolor{black}{0.9174}
					& \textcolor{blue}{33.71} & \textcolor{blue}{0.9178}
					& \textcolor{red}{33.74} & \textcolor{red}{0.9181}
					\\
					& 40
					& 34.47 & 0.9299
					& 34.51 & 0.9302
					& 34.58 & 0.9312
					& \textcolor{black}{34.67} & \textcolor{black}{0.9317}
					& \textcolor{blue}{34.70} & \textcolor{blue}{0.9322}
					& \textcolor{red}{34.73} & \textcolor{red}{0.9324}
					\\
					\hline   
					
			\end{tabular}  }
			\vspace{-4mm}
			\caption{PSNR (dB)/SSIM comparisons. The best and second best results are in \textcolor{red}{red} and \textcolor{blue}{blue}.}
			\label{table:psnr_ssim_psnrb_JPEG}
		\end{center}
		\vspace{-7mm}
	\end{table*}
	
	\noindent \textbf{Visual Comparisons.} The visual comparison for color image denoising of different methods is shown in Fig.~\ref{fig:img_dn_visual}. Our ART can preserve detailed textures and high-frequency components and remove heavy noise corruption. Compared with other methods, it has better performance to restore clean and crisp images. It demonstrates that our ART is also suitable for image denoising. 
	
	\vspace{-1.5mm}
	\subsection{JPEG Compression Artifact Reduction}
	\vspace{-2mm}
	We compare our ART with state-of-the-art JPEG CAR methods: RNAN~\cite{zhang2019rnan}, RDN~\cite{zhang2020rdnir}, DRUNet~\cite{zhang2021plugDRUNet}, and SwinIR~\cite{swinir2021}. Following most recent works, we set the compression quality factors of original images to 40, 30, and 10. We provide the PSNR and SSIM comparison results in Table~\ref{table:psnr_ssim_psnrb_JPEG}.
	
	\textbf{Quantitative Comparisons.}
	Table~\ref{table:psnr_ssim_psnrb_JPEG} shows the PSNR/SSIM comparisons of our ART with existing state-of-the-art methods. We can see that our proposed method has the best performance. Better results are achieved by ART+ using self-ensemble. These results indicate that our ART also performs outstandingly when solving image compression artifact reduction problems.
	
	\vspace{-4mm}
	\section{Conclusion}
	\vspace{-4mm}
	In this work, we propose Attention Retractable Transformer for image restoration named ART, which offers two types of self-attention blocks to enhance the Transformer representation ability. Most previous image restoration Transformer backbones mainly utilize dense attention modules to alleviate self-attention computation within non-overlapping regions and thus suffer from restricted receptive fields. Without introducing additional computational cost, we employ the sparse attention mechanism to enable tokens from sparse areas of the image to interact with each other. In practice, the alternating application of dense and sparse attention modules is able to provide retractable attention for the model and bring promising improvement. Experiments on image SR, denoising, and JPEG CAR tasks validate that our method achieves state-of-the-art results on various benchmark datasets both quantitatively and visually. In future work, we will try to apply our proposed method to more image restoration tasks, like image deraining, deblurring, dehazing, and so on. We will further explore the potential of sparse attention in solving low-level vision problems.
	
	\section*{Acknowledgments}
	This work was supported in part by NSFC grant 62141220, 61972253, U1908212, 62172276, 61972254, the Program for Professor of Special Appointment (Eastern Scholar) at Shanghai Institutions of Higher Learning, the National Natural Science Foundation of China under Grant No. 62271414, Zhejiang Provincial Natural Science Foundation of China under Grant No. LR23F010001.  This work was also supported by the Shenzhen Science and Technology Project (JCYJ20200109142808034), and in part by Guangdong Special Support (2019TX05X187). Xin Yuan would like to thank Research Center for Industries of the Future (RCIF) at Westlake University for supporting this work.

	\section*{Reproducibility Statement}
	We provide the reproducibility statement of our proposed method in this section. We introduce the model architecture and core dense and sparse attention modules in Sec.~\ref{method}. Besides, we also give the implementation details. In Sec.~\ref{subsec:settings}, we provide the detailed experiment settings. To ensure the reproducibility, we provide the source code and pre-trained models at the website\footnote{\href{https://github.com/gladzhang/ART}{https://github.com/gladzhang/ART}}. Everyone can run our code to check the training and testing process according to the given instructions. At the website, the pre-trained models are provided to verify the validity of corresponding results. More details please refer to the website or the submitted supplementary materials.
	
	\small

	\bibliography{iclr2023_conference}
	\bibliographystyle{iclr2023_conference}

\end{document}

%% file: math_commands.tex
\usepackage{amsmath,amsfonts,bm}

\def\eqref#1{equation~\ref{#1}}
\def\1{\bm{1}}

\DeclareMathAlphabet{\mathsfit}{\encodingdefault}{\sfdefault}{m}{sl}
\SetMathAlphabet{\mathsfit}{bold}{\encodingdefault}{\sfdefault}{bx}{n}

%% file: ART_arxiv.bbl
\begin{thebibliography}{75}
\providecommand{\natexlab}[1]{#1}
\providecommand{\url}[1]{\texttt{#1}}
\expandafter\ifx\csname urlstyle\endcsname\relax
  \providecommand{\doi}[1]{doi: #1}\else
  \providecommand{\doi}{doi: \begingroup \urlstyle{rm}\Url}\fi

\bibitem[Anwar \& Barnes(2020)Anwar and Barnes]{anwar2020densely}
Saeed Anwar and Nick Barnes.
\newblock Densely residual laplacian super-resolution.
\newblock \emph{TPAMI}, 2020.

\bibitem[Arbelaez et~al.(2010)Arbelaez, Maire, Fowlkes, and
  Malik]{arbelaez2010contourbsd500}
Pablo Arbelaez, Michael Maire, Charless Fowlkes, and Jitendra Malik.
\newblock Contour detection and hierarchical image segmentation.
\newblock \emph{TPAMI}, 2010.

\bibitem[Bevilacqua et~al.(2012)Bevilacqua, Roumy, Guillemot, and
  Alberi-Morel]{bevilacqua2012low}
Marco Bevilacqua, Aline Roumy, Christine Guillemot, and Marie~Line
  Alberi-Morel.
\newblock Low-complexity single-image super-resolution based on nonnegative
  neighbor embedding.
\newblock In \emph{BMVC}, 2012.

\bibitem[Charbonnier et~al.(1994)Charbonnier, Blanc-Feraud, Aubert, and
  Barlaud]{charbonnier1994two}
Pierre Charbonnier, Laure Blanc-Feraud, Gilles Aubert, and Michel Barlaud.
\newblock Two deterministic half-quadratic regularization algorithms for
  computed imaging.
\newblock In \emph{ICIP}, 1994.

\bibitem[Chen et~al.(2021{\natexlab{a}})Chen, Wang, Guo, Xu, Deng, Liu, Ma, Xu,
  Xu, and Gao]{chen2021preIPT}
Hanting Chen, Yunhe Wang, Tianyu Guo, Chang Xu, Yiping Deng, Zhenhua Liu, Siwei
  Ma, Chunjing Xu, Chao Xu, and Wen Gao.
\newblock Pre-trained image processing transformer.
\newblock In \emph{CVPR}, 2021{\natexlab{a}}.

\bibitem[Chen et~al.(2021{\natexlab{b}})Chen, Gu, and Zhang]{chen2021attention}
Haoyu Chen, Jinjin Gu, and Zhi Zhang.
\newblock Attention in attention network for image super-resolution.
\newblock \emph{arXiv preprint arXiv:2104.09497}, 2021{\natexlab{b}}.

\bibitem[Chu et~al.(2021)Chu, Tian, Wang, Zhang, Ren, Wei, Xia, and
  Shen]{chu2021twins}
Xiangxiang Chu, Zhi Tian, Yuqing Wang, Bo~Zhang, Haibing Ren, Xiaolin Wei,
  Huaxia Xia, and Chunhua Shen.
\newblock Twins: Revisiting the design of spatial attention in vision
  transformers.
\newblock In \emph{NeurIPS}, 2021.

\bibitem[Dabov et~al.(2007)Dabov, Foi, Katkovnik, and
  Egiazarian]{Dabov2007CBM3D}
Kostadin Dabov, Alessandro Foi, Vladimir Katkovnik, and Karen~O. Egiazarian.
\newblock Color image denoising via sparse 3d collaborative filtering with
  grouping constraint in luminance-chrominance space.
\newblock In \emph{ICIP}, 2007.

\bibitem[Dai et~al.(2019)Dai, Cai, Zhang, Xia, and Zhang]{dai2019secondSAN}
Tao Dai, Jianrui Cai, Yongbing Zhang, Shu-Tao Xia, and Lei Zhang.
\newblock Second-order attention network for single image super-resolution.
\newblock In \emph{CVPR}, 2019.

\bibitem[Dong et~al.(2014)Dong, Loy, He, and Tang]{dong2014learning}
Chao Dong, Chen~Change Loy, Kaiming He, and Xiaoou Tang.
\newblock Learning a deep convolutional network for image super-resolution.
\newblock In \emph{ECCV}, 2014.

\bibitem[Dong et~al.(2016)Dong, Loy, and Tang]{dong2016accelerating}
Chao Dong, Chen~Change Loy, and Xiaoou Tang.
\newblock Accelerating the super-resolution convolutional neural network.
\newblock In \emph{ECCV}, 2016.

\bibitem[Dosovitskiy et~al.(2021)Dosovitskiy, Beyer, Kolesnikov, Weissenborn,
  Zhai, Unterthiner, Dehghani, Minderer, Heigold, Gelly,
  et~al.]{dosovitskiy2020image}
Alexey Dosovitskiy, Lucas Beyer, Alexander Kolesnikov, Dirk Weissenborn,
  Xiaohua Zhai, Thomas Unterthiner, Mostafa Dehghani, Matthias Minderer, Georg
  Heigold, Sylvain Gelly, et~al.
\newblock An image is worth 16x16 words: Transformers for image recognition at
  scale.
\newblock In \emph{ICLR}, 2021.

\bibitem[Dudhane et~al.(2022)Dudhane, Zamir, Khan, Khan, and
  Yang]{dudhane2021burst}
Akshay Dudhane, Syed~Waqas Zamir, Salman Khan, Fahad Khan, and Ming-Hsuan Yang.
\newblock Burst image restoration and enhancement.
\newblock In \emph{CVPR}, 2022.

\bibitem[Foi et~al.(2007)Foi, Katkovnik, and Egiazarian]{foi2007pointwise}
Alessandro Foi, Vladimir Katkovnik, and Karen Egiazarian.
\newblock Pointwise shape-adaptive dct for high-quality denoising and
  deblocking of grayscale and color images.
\newblock \emph{TIP}, May 2007.

\bibitem[Franzen(1999)]{franzen1999kodak}
Rich Franzen.
\newblock Kodak lossless true color image suite.
\newblock \emph{source: http://r0k. us/graphics/kodak}, 1999.

\bibitem[Haris et~al.(2018)Haris, Shakhnarovich, and Ukita]{haris2018deepDBPN}
Muhammad Haris, Greg Shakhnarovich, and Norimichi Ukita.
\newblock Deep back-projection networks for super-resolution.
\newblock In \emph{CVPR}, 2018.

\bibitem[He et~al.(2010)He, Sun, and Tang]{he2010single}
Kaiming He, Jian Sun, and Xiaoou Tang.
\newblock Single image haze removal using dark channel prior.
\newblock \emph{TPAMI}, 2010.

\bibitem[Hu et~al.(2019)Hu, Zhang, Xie, and Lin]{hu2019local}
Han Hu, Zheng Zhang, Zhenda Xie, and Stephen Lin.
\newblock Local relation networks for image recognition.
\newblock In \emph{ICCV}, 2019.

\bibitem[Hu et~al.(2021)Hu, Ma, Liu, Cai, Zhao, Zhang, and Wang]{hu2021pseudo}
Xiaowan Hu, Ruijun Ma, Zhihong Liu, Yuanhao Cai, Xiaole Zhao, Yulun Zhang, and
  Haoqian Wang.
\newblock Pseudo 3d auto-correlation network for real image denoising.
\newblock In \emph{CVPR}, 2021.

\bibitem[Huang et~al.(2015)Huang, Singh, and Ahuja]{huang2015single}
Jia-Bin Huang, Abhishek Singh, and Narendra Ahuja.
\newblock Single image super-resolution from transformed self-exemplars.
\newblock In \emph{CVPR}, 2015.

\bibitem[Kim et~al.(2016{\natexlab{a}})Kim, Kwon~Lee, and
  Mu~Lee]{kim2016accurate}
Jiwon Kim, Jung Kwon~Lee, and Kyoung Mu~Lee.
\newblock Accurate image super-resolution using very deep convolutional
  networks.
\newblock In \emph{CVPR}, 2016{\natexlab{a}}.

\bibitem[Kim et~al.(2016{\natexlab{b}})Kim, Kwon~Lee, and
  Mu~Lee]{kim2016deeply}
Jiwon Kim, Jung Kwon~Lee, and Kyoung Mu~Lee.
\newblock Deeply-recursive convolutional network for image super-resolution.
\newblock In \emph{CVPR}, 2016{\natexlab{b}}.

\bibitem[Kingma \& Ba(2015)Kingma and Ba]{kingma2014adam}
Diederik Kingma and Jimmy Ba.
\newblock Adam: A method for stochastic optimization.
\newblock In \emph{ICLR}, 2015.

\bibitem[Kong et~al.(2022)Kong, Liu, Gu, Qiao, and Dong]{kong2022reflash}
Xiangtao Kong, Xina Liu, Jinjin Gu, Yu~Qiao, and Chao Dong.
\newblock Reflash dropout in image super-resolution.
\newblock In \emph{CVPR}, pp.\  6002--6012, 2022.

\bibitem[Lai et~al.(2017)Lai, Huang, Ahuja, and Yang]{lai2017deep}
Wei-Sheng Lai, Jia-Bin Huang, Narendra Ahuja, and Ming-Hsuan Yang.
\newblock Deep laplacian pyramid networks for fast and accurate
  super-resolution.
\newblock In \emph{CVPR}, 2017.

\bibitem[Li et~al.(2019)Li, Yang, Liu, Yang, Jeon, and Wu]{li2019feedbackSRFBN}
Zhen Li, Jinglei Yang, Zheng Liu, Xiaomin Yang, Gwanggil Jeon, and Wei Wu.
\newblock Feedback network for image super-resolution.
\newblock In \emph{CVPR}, 2019.

\bibitem[Li et~al.(2022)Li, Liu, Chen, Cai, Gu, Qiao, and
  Dong]{li2022blueprint}
Zheyuan Li, Yingqi Liu, Xiangyu Chen, Haoming Cai, Jinjin Gu, Yu~Qiao, and Chao
  Dong.
\newblock Blueprint separable residual network for efficient image
  super-resolution.
\newblock In \emph{CVPR}, pp.\  833--843, 2022.

\bibitem[Liang et~al.(2021)Liang, Cao, Sun, Zhang, Gool, and
  Timofte]{swinir2021}
Jingyun Liang, Jiezhang Cao, Guolei Sun, Kai Zhang, Luc~Van Gool, and Radu
  Timofte.
\newblock Swinir: Image restoration using swin transformer.
\newblock In \emph{ICCVW}, 2021.

\bibitem[Lim et~al.(2017)Lim, Son, Kim, Nah, and Lee]{lim2017enhanced}
Bee Lim, Sanghyun Son, Heewon Kim, Seungjun Nah, and Kyoung~Mu Lee.
\newblock Enhanced deep residual networks for single image super-resolution.
\newblock In \emph{CVPRW}, 2017.

\bibitem[Liu et~al.(2020)Liu, Zhang, Tang, Tang, and Wu]{liu2020residualRFANet}
Jie Liu, Wenjie Zhang, Yuting Tang, Jie Tang, and Gangshan Wu.
\newblock Residual feature aggregation network for image super-resolution.
\newblock In \emph{CVPR}, 2020.

\bibitem[Liu et~al.(2021)Liu, Lin, Cao, Hu, Wei, Zhang, Lin, and
  Guo]{swintransformer2021}
Ze~Liu, Yutong Lin, Yue Cao, Han Hu, Yixuan Wei, Zheng Zhang, Stephen Lin, and
  Baining Guo.
\newblock Swin transformer: Hierarchical vision transformer using shifted
  windows.
\newblock In \emph{ICCV}, 2021.

\bibitem[Ma et~al.(2016)Ma, Duanmu, Wu, Wang, Yong, Li, and
  Zhang]{ma2016waterloowed}
Kede Ma, Zhengfang Duanmu, Qingbo Wu, Zhou Wang, Hongwei Yong, Hongliang Li,
  and Lei Zhang.
\newblock Waterloo exploration database: New challenges for image quality
  assessment models.
\newblock \emph{TIP}, 2016.

\bibitem[Martin et~al.(2001)Martin, Fowlkes, Tal, and
  Malik]{martin2001database}
David Martin, Charless Fowlkes, Doron Tal, and Jitendra Malik.
\newblock A database of human segmented natural images and its application to
  evaluating segmentation algorithms and measuring ecological statistics.
\newblock In \emph{ICCV}, 2001.

\bibitem[Matsui et~al.(2017)Matsui, Ito, Aramaki, Fujimoto, Ogawa, Yamasaki,
  and Aizawa]{matsui2017sketch}
Yusuke Matsui, Kota Ito, Yuji Aramaki, Azuma Fujimoto, Toru Ogawa, Toshihiko
  Yamasaki, and Kiyoharu Aizawa.
\newblock Sketch-based manga retrieval using manga109 dataset.
\newblock \emph{Multimedia Tools and Applications}, 2017.

\bibitem[Mei et~al.(2020)Mei, Fan, Zhou, Huang, Huang, and
  Shi]{mei2020imageCSNLN}
Yiqun Mei, Yuchen Fan, Yuqian Zhou, Lichao Huang, Thomas~S Huang, and Humphrey
  Shi.
\newblock Image super-resolution with cross-scale non-local attention and
  exhaustive self-exemplars mining.
\newblock In \emph{CVPR}, 2020.

\bibitem[Mei et~al.(2021)Mei, Fan, and Zhou]{mei2021imageNLSA}
Yiqun Mei, Yuchen Fan, and Yuqian Zhou.
\newblock Image super-resolution with non-local sparse attention.
\newblock In \emph{CVPR}, 2021.

\bibitem[Michaeli \& Irani(2013)Michaeli and Irani]{michaeli2013nonparametric}
Tomer Michaeli and Michal Irani.
\newblock Nonparametric blind super-resolution.
\newblock In \emph{ICCV}, 2013.

\bibitem[Niu et~al.(2020)Niu, Wen, Ren, Zhang, Yang, Wang, Zhang, Cao, and
  Shen]{niu2020singleHAN}
Ben Niu, Weilei Wen, Wenqi Ren, Xiangde Zhang, Lianping Yang, Shuzhen Wang,
  Kaihao Zhang, Xiaochun Cao, and Haifeng Shen.
\newblock Single image super-resolution via a holistic attention network.
\newblock In \emph{ECCV}, 2020.

\bibitem[Paszke et~al.(2017)Paszke, Gross, Chintala, Chanan, Yang, DeVito, Lin,
  Desmaison, Antiga, and Lerer]{paszke2017automatic}
Adam Paszke, Sam Gross, Soumith Chintala, Gregory Chanan, Edward Yang, Zachary
  DeVito, Zeming Lin, Alban Desmaison, Luca Antiga, and Adam Lerer.
\newblock Automatic differentiation in pytorch.
\newblock 2017.

\bibitem[Ramachandran et~al.(2019)Ramachandran, Parmar, Vaswani, Bello,
  Levskaya, and Shlens]{ramachandran2019stand}
Prajit Ramachandran, Niki Parmar, Ashish Vaswani, Irwan Bello, Anselm Levskaya,
  and Jon Shlens.
\newblock Stand-alone self-attention in vision models.
\newblock In \emph{NeurIPS}, 2019.

\bibitem[Sajjadi et~al.(2017)Sajjadi, Sch{\"o}lkopf, and
  Hirsch]{sajjadi2017enhancenet}
Mehdi~SM Sajjadi, Bernhard Sch{\"o}lkopf, and Michael Hirsch.
\newblock Enhancenet: Single image super-resolution through automated texture
  synthesis.
\newblock In \emph{ICCV}, 2017.

\bibitem[Sheikh et~al.(2006)Sheikh, Sabir, and Bovik]{sheikh2006statistical}
Hamid~R Sheikh, Muhammad~F Sabir, and Alan~C Bovik.
\newblock A statistical evaluation of recent full reference image quality
  assessment algorithms.
\newblock \emph{TIP}, 2006.

\bibitem[Shi et~al.(2016)Shi, Caballero, Husz{\'a}r, Totz, Aitken, Bishop,
  Rueckert, and Wang]{shi2016real}
Wenzhe Shi, Jose Caballero, Ferenc Husz{\'a}r, Johannes Totz, Andrew~P Aitken,
  Rob Bishop, Daniel Rueckert, and Zehan Wang.
\newblock Real-time single image and video super-resolution using an efficient
  sub-pixel convolutional neural network.
\newblock In \emph{CVPR}, 2016.

\bibitem[Tai et~al.(2017)Tai, Yang, Liu, and Xu]{tai2017memnet}
Ying Tai, Jian Yang, Xiaoming Liu, and Chunyan Xu.
\newblock Memnet: A persistent memory network for image restoration.
\newblock In \emph{ICCV}, 2017.

\bibitem[Tian et~al.(2020)Tian, Xu, and Zuo]{tian2020imageBRDNet}
Chunwei Tian, Yong Xu, and Wangmeng Zuo.
\newblock Image denoising using deep cnn with batch renormalization.
\newblock \emph{Neural Networks}, 2020.

\bibitem[Timofte et~al.(2013)Timofte, De, and Gool]{timofte2013anchored}
Radu Timofte, Vincent De, and Luc~Van Gool.
\newblock Anchored neighborhood regression for fast example-based
  super-resolution.
\newblock In \emph{ICCV}, 2013.

\bibitem[Timofte et~al.(2017)Timofte, Agustsson, Van~Gool, Yang, Zhang, Lim,
  Son, Kim, Nah, Lee, et~al.]{timofte2017ntire}
Radu Timofte, Eirikur Agustsson, Luc Van~Gool, Ming-Hsuan Yang, Lei Zhang, Bee
  Lim, Sanghyun Son, Heewon Kim, Seungjun Nah, Kyoung~Mu Lee, et~al.
\newblock Ntire 2017 challenge on single image super-resolution: Methods and
  results.
\newblock In \emph{CVPRW}, 2017.

\bibitem[Touvron et~al.(2021)Touvron, Cord, Douze, Massa, Sablayrolles, and
  J{\'e}gou]{touvron2021training}
Hugo Touvron, Matthieu Cord, Matthijs Douze, Francisco Massa, Alexandre
  Sablayrolles, and Herv{\'e} J{\'e}gou.
\newblock Training data-efficient image transformers \& distillation through
  attention.
\newblock In \emph{ICML}, 2021.

\bibitem[Tu et~al.(2022{\natexlab{a}})Tu, Talebi, Zhang, Yang, Milanfar, Bovik,
  and Li]{tu2022maxim}
Zhengzhong Tu, Hossein Talebi, Han Zhang, Feng Yang, Peyman Milanfar, Alan
  Bovik, and Yinxiao Li.
\newblock Maxim: Multi-axis mlp for image processing.
\newblock In \emph{CVPR}, 2022{\natexlab{a}}.

\bibitem[Tu et~al.(2022{\natexlab{b}})Tu, Talebi, Zhang, Yang, Milanfar, Bovik,
  and Li]{tu2022maxvit}
Zhengzhong Tu, Hossein Talebi, Han Zhang, Feng Yang, Peyman Milanfar, Alan
  Bovik, and Yinxiao Li.
\newblock Maxvit: Multi-axis vision transformer.
\newblock In \emph{ECCV}, 2022{\natexlab{b}}.

\bibitem[Vaswani et~al.(2017)Vaswani, Shazeer, Parmar, Uszkoreit, Jones, Gomez,
  Kaiser, and Polosukhin]{vaswani2017attention}
Ashish Vaswani, Noam Shazeer, Niki Parmar, Jakob Uszkoreit, Llion Jones,
  Aidan~N Gomez, {\L}ukasz Kaiser, and Illia Polosukhin.
\newblock Attention is all you need.
\newblock In \emph{NeurIPS}, 2017.

\bibitem[Wang et~al.(2020)Wang, Zhu, Green, Adam, Yuille, and
  Chen]{wang2020axial}
Huiyu Wang, Yukun Zhu, Bradley Green, Hartwig Adam, Alan Yuille, and
  Liang-Chieh Chen.
\newblock Axial-deeplab: Stand-alone axial-attention for panoptic segmentation.
\newblock In \emph{ECCV}, 2020.

\bibitem[Wang et~al.(2021)Wang, Xie, Li, Fan, Song, Liang, Lu, Luo, and
  Shao]{pvt2021}
Wenhai Wang, Enze Xie, Xiang Li, Deng{-}Ping Fan, Kaitao Song, Ding Liang, Tong
  Lu, Ping Luo, and Ling Shao.
\newblock Pyramid vision transformer: {A} versatile backbone for dense
  prediction without convolutions.
\newblock In \emph{ICCV}, 2021.

\bibitem[Wang et~al.(2022{\natexlab{a}})Wang, Yao, Chen, Lin, Cai, He, and
  Liu]{wang2021crossformer}
Wenxiao Wang, Lu~Yao, Long Chen, Binbin Lin, Deng Cai, Xiaofei He, and Wei Liu.
\newblock Crossformer: A versatile vision transformer hinging on cross-scale
  attention.
\newblock In \emph{ICLR}, 2022{\natexlab{a}}.

\bibitem[Wang et~al.(2022{\natexlab{b}})Wang, Cun, Bao, Zhou, Liu, and
  Li]{wang2022uformer}
Zhendong Wang, Xiaodong Cun, Jianmin Bao, Wengang Zhou, Jianzhuang Liu, and
  Houqiang Li.
\newblock Uformer: A general u-shaped transformer for image restoration.
\newblock In \emph{CVPR}, 2022{\natexlab{b}}.

\bibitem[Wang et~al.(2004)Wang, Bovik, Sheikh, and Simoncelli]{wang2004image}
Zhou Wang, Alan~C Bovik, Hamid~R Sheikh, and Eero~P Simoncelli.
\newblock Image quality assessment: from error visibility to structural
  similarity.
\newblock \emph{TIP}, 2004.

\bibitem[Xia \& Chakrabarti(2020)Xia and Chakrabarti]{xia2020identifyingRPCNN}
Zhihao Xia and Ayan Chakrabarti.
\newblock Identifying recurring patterns with deep neural networks for natural
  image denoising.
\newblock In \emph{WACV}, 2020.

\bibitem[Yang et~al.(2020)Yang, Yang, Fu, Lu, and Guo]{yang2020learning}
Fuzhi Yang, Huan Yang, Jianlong Fu, Hongtao Lu, and Baining Guo.
\newblock Learning texture transformer network for image super-resolution.
\newblock In \emph{CVPR}, 2020.

\bibitem[Yu et~al.(2021)Yu, Xia, Bai, Lu, Yuille, and Shen]{gg-transformer2021}
Qihang Yu, Yingda Xia, Yutong Bai, Yongyi Lu, Alan~L Yuille, and Wei Shen.
\newblock Glance-and-gaze vision transformer.
\newblock In \emph{NeurIPS}, 2021.

\bibitem[Zamir et~al.(2020)Zamir, Arora, Khan, Hayat, Khan, Yang, and
  Shao]{zamir2020learning}
Syed~Waqas Zamir, Aditya Arora, Salman Khan, Munawar Hayat, Fahad~Shahbaz Khan,
  Ming-Hsuan Yang, and Ling Shao.
\newblock Learning enriched features for real image restoration and
  enhancement.
\newblock In \emph{ECCV}, 2020.

\bibitem[Zamir et~al.(2021)Zamir, Arora, Khan, Hayat, Khan, Yang, and
  Shao]{zamir2021multi}
Syed~Waqas Zamir, Aditya Arora, Salman Khan, Munawar Hayat, Fahad~Shahbaz Khan,
  Ming-Hsuan Yang, and Ling Shao.
\newblock Multi-stage progressive image restoration.
\newblock In \emph{CVPR}, 2021.

\bibitem[Zamir et~al.(2022)Zamir, Arora, Khan, Hayat, Khan, and
  Yang]{restormer2022}
Syed~Waqas Zamir, Aditya Arora, Salman~H. Khan, Munawar Hayat, Fahad~Shahbaz
  Khan, and Ming{-}Hsuan Yang.
\newblock Restormer: Efficient transformer for high-resolution image
  restoration.
\newblock In \emph{CVPR}, 2022.

\bibitem[Zeyde et~al.(2010)Zeyde, Elad, and Protter]{zeyde2012single}
Roman Zeyde, Michael Elad, and Matan Protter.
\newblock On single image scale-up using sparse-representations.
\newblock In \emph{Proc. 7th Int. Conf. Curves Surf.}, 2010.

\bibitem[Zhang et~al.(2017{\natexlab{a}})Zhang, Zuo, Chen, Meng, and
  Zhang]{zhang2017beyonddncnn}
Kai Zhang, Wangmeng Zuo, Yunjin Chen, Deyu Meng, and Lei Zhang.
\newblock Beyond a gaussian denoiser: Residual learning of deep cnn for image
  denoising.
\newblock \emph{TIP}, 2017{\natexlab{a}}.

\bibitem[Zhang et~al.(2017{\natexlab{b}})Zhang, Zuo, Gu, and
  Zhang]{zhang2017learningIRCNN}
Kai Zhang, Wangmeng Zuo, Shuhang Gu, and Lei Zhang.
\newblock Learning deep cnn denoiser prior for image restoration.
\newblock In \emph{CVPR}, 2017{\natexlab{b}}.

\bibitem[Zhang et~al.(2018{\natexlab{a}})Zhang, Zuo, and
  Zhang]{zhang2018ffdnet}
Kai Zhang, Wangmeng Zuo, and Lei Zhang.
\newblock Ffdnet: Toward a fast and flexible solution for cnn-based image
  denoising.
\newblock \emph{TIP}, 2018{\natexlab{a}}.

\bibitem[Zhang et~al.(2021{\natexlab{a}})Zhang, Li, Zuo, Zhang, Van~Gool, and
  Timofte]{zhang2021plugDRUNet}
Kai Zhang, Yawei Li, Wangmeng Zuo, Lei Zhang, Luc Van~Gool, and Radu Timofte.
\newblock Plug-and-play image restoration with deep denoiser prior.
\newblock \emph{TPAMI}, 2021{\natexlab{a}}.

\bibitem[Zhang et~al.(2011)Zhang, Wu, Buades, and Li]{zhang2011color}
Lei Zhang, Xiaolin Wu, Antoni Buades, and Xin Li.
\newblock Color demosaicking by local directional interpolation and nonlocal
  adaptive thresholding.
\newblock \emph{J Electron Imaging}, 2011.

\bibitem[Zhang et~al.(2018{\natexlab{b}})Zhang, Li, Li, Wang, Zhong, and
  Fu]{zhang2018image}
Yulun Zhang, Kunpeng Li, Kai Li, Lichen Wang, Bineng Zhong, and Yun Fu.
\newblock Image super-resolution using very deep residual channel attention
  networks.
\newblock In \emph{ECCV}, 2018{\natexlab{b}}.

\bibitem[Zhang et~al.(2019)Zhang, Li, Li, Zhong, and Fu]{zhang2019rnan}
Yulun Zhang, Kunpeng Li, Kai Li, Bineng Zhong, and Yun Fu.
\newblock Residual non-local attention networks for image restoration.
\newblock In \emph{ICLR}, 2019.

\bibitem[Zhang et~al.(2020)Zhang, Tian, Kong, Zhong, and Fu]{zhang2020rdnir}
Yulun Zhang, Yapeng Tian, Yu~Kong, Bineng Zhong, and Yun Fu.
\newblock Residual dense network for image restoration.
\newblock \emph{TPAMI}, 2020.

\bibitem[Zhang et~al.(2021{\natexlab{b}})Zhang, Wang, Qin, and Fu]{zhangASSL}
Yulun Zhang, Huan Wang, Can Qin, and Yun Fu.
\newblock Aligned structured sparsity learning for efficient image
  super-resolution.
\newblock In \emph{NeurIPS}, 2021{\natexlab{b}}.

\bibitem[Zhao et~al.(2020)Zhao, Jia, and Koltun]{zhao2020exploring}
Hengshuang Zhao, Jiaya Jia, and Vladlen Koltun.
\newblock Exploring self-attention for image recognition.
\newblock In \emph{CVPR}, 2020.

\bibitem[Zheng et~al.(2021)Zheng, Lu, Zhao, Zhu, Luo, Wang, Fu, Feng, Xiang,
  Torr, et~al.]{zheng2021rethinking}
Sixiao Zheng, Jiachen Lu, Hengshuang Zhao, Xiatian Zhu, Zekun Luo, Yabiao Wang,
  Yanwei Fu, Jianfeng Feng, Tao Xiang, Philip~HS Torr, et~al.
\newblock Rethinking semantic segmentation from a sequence-to-sequence
  perspective with transformers.
\newblock In \emph{CVPR}, 2021.

\bibitem[Zhou et~al.(2020)Zhou, Zhang, Zuo, and Loy]{zhou2020crossIGNN}
Shangchen Zhou, Jiawei Zhang, Wangmeng Zuo, and Chen~Change Loy.
\newblock Cross-scale internal graph neural network for image super-resolution.
\newblock In \emph{NeurIPS}, 2020.

\end{thebibliography}
